\pdfoutput=1
\documentclass{bmvc2k}

\usepackage{hyperref}

\usepackage{algorithm}
\usepackage{algorithmic}
\usepackage{amsmath}
\usepackage{graphicx}
\usepackage{multirow}
\usepackage{array}

\usepackage{xspace}

\providecommand{\figref}{}
\renewcommand{\figref}[1]{Fig.~\ref{#1}}
\newcommand{\tabref}[1]{Tab.~\ref{#1}}
\providecommand{\secref}{}
\renewcommand{\secref}[1]{\S\ref{#1}}
\providecommand{\secrefs}{}
\renewcommand{\secrefs}[2]{\S\ref{#1}--\ref{#2}}
\providecommand{\appref}{}
\renewcommand{\appref}[1]{App.~\ref{#1}}
\newcommand{\algoref}[1]{Alg.~\ref{#1}}

\newcommand{\parheading}[1]{\smallskip{}\noindent{}\textbf{#1}\hspace{2pt}}

\newcommand{\facerec}[0]{FR\xspace}
\newcommand{\fr}[0]{\facerec}
\newcommand{\ml}[0]{ML\xspace}
\newcommand{\pgd}[0]{PGD\xspace}
\newcommand{\fgsm}[0]{FGSM\xspace}
\DeclareRobustCommand{\attack}{\textsc{BoundStyle}\xspace}
\DeclareRobustCommand{\defense}{\textsc{StyleAT}\xspace}
\newcommand{\lpnorm}[1]{\ensuremath{\ell_{#1}}\xspace}
\newcommand{\diffprivate}[0]{DiffPrivate\xspace}
\newcommand{\DOA}[0]{DOA\xspace}
\newcommand{\diffpure}[0]{DiffPure\xspace}

\usepackage{booktabs,subcaption}

\addauthor{Ben Shapira}{benshapira@mail.tau.ac.il}{1}
\addauthor{Roi Cohen}{Roi.Cohen@hpi.de}{2}
\addauthor{Shang-Tse Chen}{stchen@csie.ntu.edu.tw}{3}
\addauthor{Mahmood Sharif}{mahmoods@tauex.tau.ac.il}{1}

\addinstitution{
 Tel Aviv University\\
 Tel Aviv, Israel
}
\addinstitution{
 HPI / University of Potsdam\\
 Potsdam, Germany
}
\addinstitution{
 National Taiwan University\\
 Taipei, Taiwan
}

\title{\defense{}: Defending Face Recognition Against Semantic Attacks}
\runninghead{Shapira et al.}{Defending Face Recognition Against Semantic Attacks}

\pdfpagewidth=\paperwidth
\begin{document}

\maketitle
\begin{abstract}

With face-recognition models now embedded in everyday authentication and surveillance, recent works have pinpointed a critical weakness: these models remain acutely vulnerable to adversarial semantic edits. 
I.e., adversarially produced semantic alterations to the input, such as slight aging or pose changes, can induce misclassifications.
Certain existing attacks are powerful, but they can be computationally costly, rendering them inadequate for developing defenses (e.g., through adversarial training). 
To fill the gap, we introduce \textbf{\attack{}}, a potent semantic attack operating in StyleGAN's rich latent space to maximize misclassification rates. 
Notably, \attack{} achieves high attack success rates while being ${\sim}{\times}9.5$ faster than existing state-of-the-art attacks, making it suitable for adversarial training. 
Building on \attack{}, we develop \textbf{\defense{}}, an efficient adversarial training scheme that incorporates low-budget attack variants yet defends against stronger and unseen semantic attacks.
We evaluate on two datasets unseen during training 
and seven models, and find that \defense{} boosts robust accuracy against state-of-the-art attacks
and outperforms common defenses 
in various settings.

\end{abstract}

\section{Introduction}

Face-recognition (\facerec{}) technologies are employed in various
security-critical applications, including for surveillance and access
control~\citep{introna2009facial}.
Failures of such systems may be pernicious; for instance, false
negatives may enable criminals to avoid surveillance, whereas false
positives may provision unauthorized access to important resources
protected by access control.
However, unfortunately, similar to other machine-learning (\ml{})
models that can be evaded by adversarial examples at inference
time~\citep{goodfellow2014explaining, szegedy2013intriguing}, \facerec{} models are vulnerable to \emph{general
semantic attacks}---a class of adversarial example attacks
that introduce slight semantic changes (e.g., addition of accessories
or alterations of pose, expression, or age) to fool \facerec{} despite
preserving the identity of the subject in the image~\citep{barattin2023attribute, jia2022adv, le2024styleadv, le2025diffprivate, liu2024adv}.

Existing general semantic attacks mostly rely on generative models,
such as StyleGAN~\citep{le2024styleadv} and latent diffusion
models~\citep{le2025diffprivate}, to discover adversarial semantic edits through
latent-space perturbations that mislead \facerec{}.
However, these attacks suffer from certain limitations:
Some attacks such as StyleAdv achieve \emph{limited attack success}~\citep{le2025diffprivate};
other attacks such as AMT-GAN produce edits with \emph{low visual
fidelity}~\citep{le2024styleadv}; and
attacks such as \diffprivate{} are computationally heavy,
requiring significant run time to attain high success rates (see \secref{sec:exp}).
Due to these limitations, existing attacks may fail to uncover
weakness in \facerec{} or may not lend themselves to
being incorporated in training schemes for improving \facerec{}'s
robustness.

For specific forms of adversarial examples, such as those created by
adversarial perturbations with bounded \lpnorm{p}-norms, numerous
defense types like adversarial training~\citep{wong2020fast} and randomized smoothing~\citep{Cohen19RandSmooth}
can help boost robustness.
However, to our knowledge, no established defenses have been
proposed to mitigate general semantic attacks. 
Particularly, adversarial training for defending against general
semantic attacks remains infeasible due to the run-time overhead or
limited success rates of existing attacks.
Consequently, \facerec{} remains vulnerable to general semantic
attacks.

To fill these gaps, this work presents a new general semantic attack,
\attack{}, and a defense, \defense{}.
Our attack takes advantage of StyleGAN3's rich latent space~\cite{karras2021alias}, among
others, to produce high-fidelity semantic edits to fool \facerec{}.
Importantly, \attack{} is tunable, enabling us to control the
magnitude of edits and run time, thus ensuring that identity is
preserved w.r.t.\ human observer and allowing us to execute
time-efficient variants.
Notably, we also find that \attack{} is highly successful,
while being significantly faster (roughly $\times$9.5) than the state-of-the-art attack, \diffprivate{}~\citep{le2025diffprivate}.

Interestingly, we find compelling evidence that \diffprivate{} introduces imperceptible adversarial perturbations alongside visible semantic edits (\appref{imperceptible_perturbations}); if used for adversarial training, models may learn to resist pixel noise rather than semantic changes. This motivates \attack{}'s design to operate purely in StyleGAN's semantic latent space. Supporting this design choice, we find no clear signs that \attack{} makes edits other than semantic ones~(\secref{subsec:defenceeval} consolidates the supporting evidence).
Altogether, \attack{}'s advantages render it suitable for 
measuring the susceptibility of \facerec{} models to attacks as well
as for adversarial training to help improve robustness against
semantic attacks.

Our defense, \defense{}, employs a time-efficient variant of \attack{}
to adversarially train \facerec{} models and improve their adversarial
robustness against general semantic attacks.
Against \attack{}, \defense{} achieves up to 28.6\%
increase in robust accuracy (depending on the setting explored)
compared to undefended models, markedly higher than defenses not tailored for general semantic attacks that achieve $\le$6.0\% increase in robust accuracy.
Crucially, \defense{} also leads to improvements against \diffprivate{}, an attack not encountered during training, with up to 46.3\% higher robust accuracy than undefended models, showcasing that \defense{} generalizes to unknown general semantic attacks.

We next present related work (\secref{sec:related}) and our threat
model (\secref{sec:threat}). Subsequently, we describe the technical approach
behind \attack{} and \defense{} (\secref{sec:methods}) before presenting our
experimental results (\secrefs{subsec:setup}{sec:exp}) and concluding (\secref{sec:conclusion}).

\section{Related Work}
\label{sec:related}


\parheading{Attacks on \facerec{}}
Prior work has demonstrated that \ml{} models in general, and \facerec{} in
particular, are vulnerable to test-time evasion attacks that induce
misclassifications via imperceptible adversarial perturbations with
bounded \lpnorm{p}-norm~\citep{szegedy2013intriguing}.
For instance, the fast gradient sign method (\fgsm{}) creates attacks
by perturbing inputs in the gradient direction once~\citep{goodfellow2014explaining}, while
projected gradient descent (\pgd{}) does so through multiple,
iterative perturbations~\cite{madry2017towards}.
However, such attacks may be challenging to realize in real-world
settings due to difficulties in implementing norm-bounded noise and
cameras' sampling errors, among others~\citep{sharif2016accessorize}.
To this end, researchers have proposed semantic attacks---attacks that
alter inputs in minor, easy-to-realize, and semantically meaningful
ways---to mislead \facerec{} models.

Semantic attacks consist of two families.
The first family of attacks makes \emph{ad hoc} changes to inputs, for example, by 
introducing adversarial accessories like eyeglasses or hats
to fool models~\citep{komkov2021advhat, sharif2016accessorize}.
These also include attacks that fool models via facial make-up
or spatial transformations applied in an adversarial
manner~\citep{hu2022protecting, xiao2018spatially, yin2021adv}.
By contrast, the second family of attacks leverages \emph{general}
edits of inputs to induce misclassifications, including, but not
limited to, changes of
expression, age, and
accessories, or a combination thereof~\citep{barattin2023attribute, jia2022adv, le2024styleadv, le2025diffprivate, liu2024adv}.
Our work focuses on general semantic attacks, proposing a new attack
and a defense.

General semantic attacks typically leverage generative models to
produce adversarial edits of inputs.
For instance, attacks such as StyleAdv~\citep{le2024styleadv} and
Adv-Attribute~\citep{jia2022adv} search for adversarial editing directions
in the latent space of generative adversarial networks (GANs) to produce misclassifications.
By contrast, Adv-Diffusion~\citep{liu2024adv} and
\diffprivate{}~\citep{le2025diffprivate} use latent diffusion models to find
adversarial semantic edits of inputs.
\diffprivate{} is the most recent and potent general semantic attack;
we use it in our evaluation.


\parheading{Defending \fr{}}
A diversity of defenses against evasion attacks have been proposed,
including, but not limited to, ones that   
detect attacks (e.g.,~\cite{Metzen17Detector}); 
filter out adversarial perturbations (e.g.,~\cite{Xu18Squeeze});
smoothen classification boundaries to reduce model vulnerability
(e.g.,~\cite{Carlini23Certified, Cohen19RandSmooth}); 
verify robustness against specific adversaries (e.g.,~\cite{Katz19Marabou}); and
adversarial train 
of models by injecting correctly labeled adversarial
inputs to the training data to inherently increase model robustness
(e.g.,~\cite{kurakin2017adversarial, madry2017towards}). 
Due to its intuitive nature, its ability to improve adversarial
robustness in a practical manner against different attack types, and
absence of impact on model's inference time,
Adversarial training is particularly appealing and was widely
studied.
Still, adversarial training may be computationally expensive due
to the overhead of producing attacks during training, potentially
rendering training prohibitive.
To this end, researchers have also explored efficient adversarial training
variants (e.g.,~\cite{Shafahi19PGD, wong2020fast}).
We take inspiration from \citet{wong2020fast} who showed
how to leverage the efficient \fgsm{} attack in training to induce
robustness against much more potent attacks at test time.

To the best of our knowledge, there are \emph{no established defenses
for countering general semantic attacks against \facerec{}}.
Nonetheless, several countermeasures have been proposed to counter ad
hoc semantic attacks.
For example, defense through occlusion attack (\DOA) adversarially
trains models with carefully positioned patches containing adversarial
patterns to help counter eyeglass attacks~\citep{wu2019defending}.
As another example, \diffpure{} utilizes forward diffusion followed by
image recovery to remove adversarial manipulations, helping counter
spatial adversarial modifications~\citep{nie2022diffusion}.
Prior work has also shown that input filters, such as JPEG
compression,  blurring, and Gaussian noise
can hinder general semantic
attacks to some degree when attacks are agnostic to the
filters~\citep{le2025diffprivate}.
Our evaluation (\secref{sec:exp}) shows that these approaches yield limited improvements in robustness; e.g., \diffpure{} proves effective against \diffprivate{} but is counterproductive against \attack{}, illustrating that the effectiveness of purification-based defenses can be attack-type dependent.
Related to our work, \citet{laidlaw2020perceptual} adversarially train \ml{} models with imperceptible adversarial perturbations created using generative models.
However, they only evaluate robustness against imperceptible perturbations and spatial manipulations.

\parheading{Relation to Deepfake Detection}
As adversarial semantic edits are synthetically produced, deepfake
detection~\cite{deepfakesurvey} may seem like a natural countermeasure.
However, detection and robust recognition solve different,
complementary problems: a detector flags whether the
image synthetic or edited, whereas \facerec{} must decide
\emph{identity}---i.e., whether a face matches the enrolled subject despite the
edit.
Even a perfect detector leaves the identity decision open; moreover,
as edits may be benign (e.g., beauty filters),
rejecting all flagged images would impose a high false-positive burden
on legitimate users.
Detection alone may also be insufficient: detectors often generalize
poorly to unseen generative models~\citep{corvi2023detection,
ojha2023universal} and can be evaded by white-box adversaries such as
ours~\citep{carlini2020evading}.
Accordingly, we view detection as a complementary defense layer, while
\defense{} hardens the component selecting identity.

\section{Threat Model}\label{sec:threat}

We consider an adversary carrying out a general semantic attack
against \facerec{}.
Per standard practice, we assume the \facerec{} system is tuned to
an operating point where the false positive rate (FPR) is below a
target threshold, such as 0.01 FPR~\citep{introna2009facial, le2025diffprivate}.
We consider a powerful adversary aiming to produce arbitrary,
untargeted misclassifications (primarily, false negatives) rather than
impersonations (i.e., targeted attacks), as, intuitively, this
adversary would be more challenging to defend against.
Contrastively, our defender aims to hinder the adversary's attempts
through keeping the robust true positive rate (TPR)---i.e., the TPR
under attacks---high while preserving the benign accuracy of standard
\facerec{} models when ingesting clean inputs.
In line with~\citet{le2025diffprivate}, We primarily focus on potent adversaries with white-box access to both
the \facerec{} model and the defense, but we also consider
black-box adversaries without access to the model or
defense that seek to transfer attacks from surrogate
models, as well as
gray-box adversaries that have access to the (undefended)
model but not to the defense.

We assume a fully digital threat model where the adversary modifies a digital image and submits it electronically. Physical re-capture scenarios (e.g., adversarial accessories worn in front of a camera) are outside scope, consistent with the digital-first framing of \diffprivate{}~\citep{le2025diffprivate} and related semantic attacks.

Our threat model captures settings where images may be
manipulated digitally or physically to mislead \facerec{}.
In cases where images arrive from devices the \facerec{} operator does
not control, such as uploads to social media or electronic
know-your-customer onboarding,
attacks may alter images to gain privacy or impersonate others.
Moreover, in idealized settings without sampling noise, semantic
attacks' edits encompass changes that may be applied physically (e.g.,
make-up or pose changes) to fool \facerec{}, providing an upper bound on
attack success in the physical domain.

\section{Technical Approach}
\label{sec:methods}

\subsection{\attack{}: A High-Fidelity, Potent, Tunable General Semantic Attack}

We design \attack{} as a general semantic attack against \fr{} based
on generative models with three goals in mind: 
\emph{(1)} We require that the attack produces evasive face images
with \emph{high visual fidelity} through diverse and realistic
edits;
\emph{(2)} We seek \emph{tunability} such that we would be able to
control the run time of the attack to later enable efficient
adversarial training and bound the magnitude of the edit so as the
identity in the face image is unchanged (for a human observer); and
\emph{(3)} We need the attack to be \emph{potent} exposing the
weaknesses of \fr{} through achieving high success rates.
We next describe how our design of \attack{} ensures high visual
fidelity and tunability.
Our experiments (\secrefs{subsec:setup}{sec:exp}) evidence the attack's potency.

\parheading{A Tunable Attack}
Let $F$ denote the feature extractor used for \fr{},
$C$ the preprocessing algorithm (cropping and alignment),
$G$ a generative model,
$x$ the image to modify with an inverted latent code $l$, and
$x_e$ a face image of the same subject enrolled in the gallery.
\attack{} aims to edit $x$ through a slight modification of $l$ such
that the similarity ($\text{sim}$, usually cosine similarity)
with $x_e$ would become small.
Formally, \attack{} aims to minimize the following loss through a
perturbation $\delta$ of the latent code:
\[
L_{\text{atk}} = \text{sim}\big(F(C(G(l+\delta))),\; F(C(x_e))\big).
\]

\attack{} optimizes the loss through iterative gradient-based
optimization, in the spirit of \pgd{}, and its performance is governed
by two primary inputs $T$, the number of iterations, and $\beta$, the
magnitude (specifically, \lpnorm{2}-norm) of the perturbation
$\delta$.
Initially, $\delta_0$ is randomly initialized inside the $\beta$-ball,
as random initialization is critical to the performance of evasion
attacks in adversarial training~\citep{wong2020fast}.
Subsequently, in each iteration (up to $T$), \attack{} updates
$\delta_i = \delta_i - \alpha \cdot \frac{g}{\|g\|_{\infty}}$
where $g = \nabla_{\delta_i} L_{atk}$ is the loss gradient and
$\alpha$ is a step-size hyperparameter.
At any point, if the norm of $\delta_i$
exceeds the bound $\beta$, it is projected back to the $\beta$-ball
via
$\delta_i = \beta \cdot \frac{\delta_i}{\|\delta_i\|_2}$.

Both $T$ and $\beta$ are tunable parameters that enable achieving
different trade-offs with \attack{}. Decreasing $T$ may potentially
harm the attack success, but also makes \attack{} faster,
rendering it more suitable for (efficient) adversarial training.
Moreover, setting $\beta$ should balance two goals---it should be
large enough so that the attack is successful due to stronger edits in
the latent space, but small enough so that the identity of the subject
is preserved w.r.t.\ human observers.

\parheading{Ensuring High Fidelity}
We take several measures to ascertain that \attack{} introduces
high-fidelity edits.
First, we adopt the StyleGAN3 generator~\citep{karras2021alias}, which
provides a rich latent space with diverse editing directions and high
quality outputs.
Second, we invert $x$ to a latent code $l$ that maps back almost
precisely to $x$, thus preserving identity. To do so, we use a
hybrid combination of encoder-based projection to the latent
space~\citep{alaluf2201third} followed by direct gradient-based
optimization for accurate reconstruction of the face
image~\citep{zhu2020domain}.
Last, we use pivotal tuning,
a method that tunes the generator $G$ to enable better editability
while preserving identity~\citep{roich2022pivotal}.

\subsection{\defense{}: Style-Aware Adversarial Training}

\begin{figure*}[t!]
    \centering
    \includegraphics[width=\textwidth]{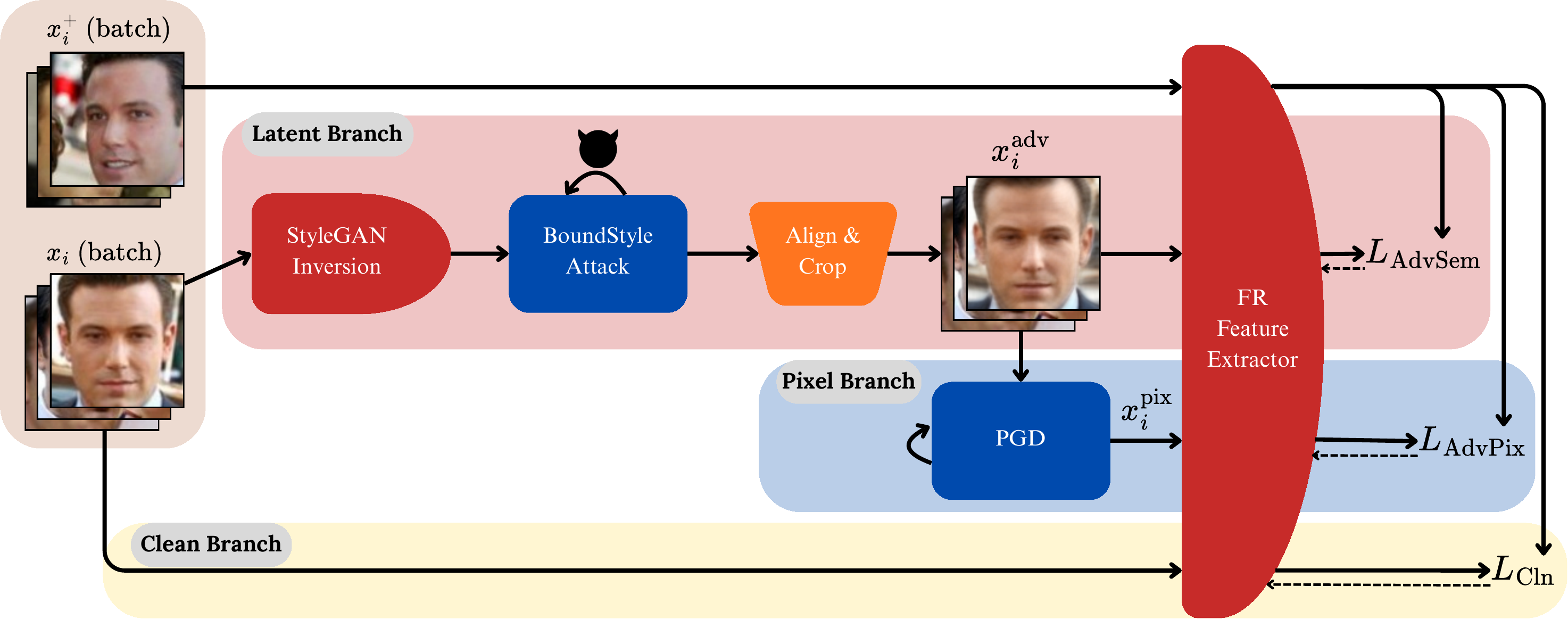}
\caption{\textbf{An overview of \defense{}'s training pipeline.}
    The framework processes a mini-batch of identity pairs $\{(x_i,
    x_i^+)\}$ through three parallel branches: (1) The \emph{clean
      branch} extracts features from the original samples to preserve
    benign accuracy ($L_{\text{Cln}}$). (2) The \emph{latent branch}
    generates a batch of semantic adversarial examples
    $x_i^{\text{adv}}$ via \attack{} to improve robustness against
    semantic attacks ($L_{\text{AdvSem}}$). (3) The
    \emph{pixel branch} applies imperceptible perturbations
    to the semantically edited batch to increase robustness against
    imperceptible perturbations ($L_{\text{AdvPix}}$). Batch
    processing enables hard negative mining to optimize the triplet
    losses.}
\label{fig:styleat_diagram}
\end{figure*}

Building on \attack{}, we design \defense{}, a method for
adversarially training \facerec{} models to enhance their adversarial
robustness against general semantic attacks.
In particular, \defense{} fine-tunes pre-trained \facerec{} feature
extractors while aiming to balance three different objectives:
\emph{(1)} Preserving benign accuracy on clean images;
\emph{(2)} Improving robust accuracy against general semantic attacks;
and 
\emph{(3)} Countering imperceptible perturbations inadvertently
introduced by certain established semantic attacks.
To achieve each of these goals, \defense{} minimizes the triplet
losses $L_\text{Cln}$, $L_\text{AdvSem}$, and
$L_\text{AdvPix}$,
respectively.
These losses are balanced through non-negative hyperparameters that
accumulate to one (i.e.,
$\lambda_\text{Cln} + \lambda_\text{AdvSem} + \lambda_\text{AdvPix} = 1$).
We next elaborate how each of $L_\text{AdvSem}$ and $L_\text{AdvPix}$
are computed and optimized and how we select triplets for loss computation;
minimizing $L_\text{Cln}$ is intuitive and follows standard
practice~\citep{wang2021facex}. An overview of \defense{}'s pipeline is depicted in \figref{fig:styleat_diagram};
\algoref{alg:style_aware_at} in \appref{app:defense_alg}
presents its pseudocode.

\parheading{Computing and Optimizing $L_\text{AdvSem}$}
We leverage \attack{} to produce general semantic attacks for
adversarial training.
However, as executing the most potent attack variant during training
may make the training process infeasible, we incorporate ``weakened''
but efficient attack variants into training.
Specifically, we run fast variants of \attack{} with a small number of
iterations $T$, analogously to fast adversarial training with
\fgsm{}~\citep{wong2020fast}.
Here, we tune $T$ and the step size $\alpha$ such that training can be
completed within a few days
under our resource constraints, while the
attacks are sufficiently evasive to help enhance \facerec{}'s
robustness against general semantic attacks.

\parheading{Computing and Optimizing $L_\text{AdvPix}$}
Our evaluation of established semantic attacks shows that certain
attacks may introduce imperceptible perturbations alongside semantic
edits to mislead \facerec{}.
This phenomenon is perhaps most clearly demonstrated when evaluating
attacks against filter-based defenses such as JPEG compression that
primarily affect imperceptible, non-semantic perturbations.
Said differently, if such filters have a pronounced effect on an
attack's success, one may conclude that misclassification did not
occur due to a semantic edit, but rather due to imperceptible changes
of pixels.
Indeed, our evaluation shows that
\diffprivate{}
exhibits a significant decrease in their success once JPEG compression
and similar filters are employed (see
\figref{fig:gbox-defenses-diffpriv} and
  \figref{fig:gbox-defenses-diffpriv-repvgg}, with \appref{imperceptible_perturbations}
  providing additional support through a frequency-domain analysis).
To account for these potential perturbations, we also adversarially
train our models against imperceptible \lpnorm{\infty}-norm-bounded
adversarial perturbations.
We find that doing so does not harm robustness against semantic attacks 
that do not seem to introduce imperceptible adversarial
perturbations (namely, \attack{}; \appref{app:ablation}). 

Toward countering imperceptible adversarial perturbations, we
integrate fast \pgd{} attacks with few iterations into training, in the spirit of Wong et al.'s (\citeyear{wong2020fast}) work.
Importantly, to avoid robust overfitting, we perform \pgd{} with
random initialization before each attack.
Crucially, we apply \pgd{} to images already edited adversarially
with \attack{}, as we aim to counter the combination of adversarial
semantic edits and imperceptible perturbations.

\parheading{Selection of Triplets}
For a given positive pair of samples depicting the same identity, we
select the hardest negative sample from the batch to compute triplet
losses.
Doing so, as shown in prior work on adversarially robust metric
learning~\citep{mao2019metriclearningadversarialrobustness}, is most conducive for adversarial robustness.
More precisely, to compute the triplet loss, for a positive pair of samples $p$ and $a$ (standing
for positive and anchor, respectively), we select the negative sample $n$ from the batch
such that it depicts a different identity and is most similar to $a$ in the
feature space compared to other samples in the batch.
Subsequently, the triplet loss is calculated by
$\left(\text{sim}(F(C(n)), F(C(a))) - \text{sim}(F(C(p)), F(C(a))) + \mu\right)^+$,
where $\mu$ is a small (non-negative) constant.
Moreover, in the interest of improving adversarial
robustness, we adversarially perturb the anchor sample
$a$ when computing $L_\text{AdvSem}$ and $L_\text{AdvPix}$, and select
the hardest negative sample after applying the perturbations, per \citet{li2019improvingrobustnessdeepneural}.

\section{Experimental Setup}
\label{subsec:setup}

\parheading{FR Backbones}
We employ seven popular and high-performing \facerec{} models, four
convolutional networks, one vision transformer, and two convolutional networks equipped with specialized supervisory heads.
Specifically, we use MobileFaceNet (MobileFace)
~\citep{chen2018mobilefacenets}; 
ResNet (ResNet-152, IR-SE) \citep{he2016deep};
RepVGG \citep{ding2021repvgg};
LightCNN \citep{wu2018light, wu2020learning}; Swin Transformer (SwinT)
\citep{liu2021swin};
and ArcFace~\citep{deng2019arcface} as well
  as MagFace~\citep{meng2021magface}
  with MobileFace backbones.
We use these models in two roles, both as targets for attacks, and as
surrogates (i.e., proxies) for producing transferable adversarial examples.
As part of \defense{}, we create adversarially trained variants of ResNet and RepVGG through fine-tuning the original pre-trained backbones.
We obtain the initial weights from FaceX-Zoo \citep{wang2021facex}.

\parheading{Datasets}
Following FaceX-Zoo \citep{wang2021facex}, we construct our training
dataset from MS-Celeb-1M-v1c~\citep{guo2016ms-celeb-}, using their randomly selected
preprocessed positive image pairs for training.
We then run our preprocessing pipeline on all images and discard pairs
where an image fails face or landmark detection,
yielding a final training set of 52{,}269 positive image pairs. (Note
that negative images are selected as hard negatives, independently for
each batch during training.)
We evaluate on two prominent datasets:
\emph{(1)} Labeled Faces in the Wild (LFW) \citep{huang2008labeled}
and
\emph{(2)} VGG-Face \citep{parkhi2015deep}.
Specifically, we select 216 and 156 positive pairs
from LFW and VGG-Face, respectively.
When selecting images, we ensure no overlap between
the selected identities and those appearing in the training (and pre-training) sets from MS-Celeb-1M-v1c,
by removing any image that has similarity with any training samples
above the threshold where the original ResNet backbone has an FPR
of $10^{-4}$.

\parheading{Attacks}
We evaluate \attack{}, bounding the edit perturbation \lpnorm{2}-norms
in StyleGAN3's latent space to $\beta\in\{1.0, 1.5, 2.0, 3.0\}$.
We avoid perturbations of larger magnitude to help preserve subject
identities in images (\appref{app:user_study}).
For best performance, we set the step size $\alpha$=$\beta$, and the number of
iterations $T$=30, as more iterations show no improvements in attack
success (\appref{ablation_iterations}).
As a baseline, we evaluate \diffprivate{} \citep{le2025diffprivate}, a
state-of-the-art attack that edits images through perturbations in the
diffusion latent $z$-space. 
For a fair comparison with \attack{}, we adopt a norm-bounded variant
of \diffprivate{} by enforcing 
$\|\Delta z\|_2=\|z_{\mathrm{adv}} - z_0\|_2 \in \{1,2,3,4,5,6\}$.
We avoid perturbations with norm $>$6 to preserve subject identities
in images (\appref{app:user_study}).
Under this bounded setup, we find that capping the optimization at 70 iterations for convergence to the highest attack success (\appref{ablation_iterations}).

\parheading{Defenses}
We apply \defense{} on both ResNet and RepVGG models, adversarially training them from checkpoints obtained from FaceX-Zoo.
We use a low-cost variant of \attack{} for adversarial training, with
$T=3$ attack iterations, $\beta=3$ \lpnorm{2}-norm for perturbations
in the StyleGAN3 latent space, and $\alpha=3$ as step size in the attack;
we run \pgd{} attacks for 2 iterations with $\epsilon=24/255$ \lpnorm{\infty}-norm for perturbations
in the pixel space and step size $\alpha=\epsilon/2$.
We find that these parameters help attain reasonable benign accuracy within feasible time under our resource constraints.
After hyperparameter search, we set specific loss weights for each backbone:
for ResNet, we set $\lambda_\text{Cln}=0.35$, $\lambda_\text{AdvSem}=0.45$, and $\lambda_\text{AdvPix}=0.2$;
for RepVGG, we set $\lambda_\text{Cln}=0.1$, $\lambda_\text{AdvSem}=0.8$, and $\lambda_\text{AdvPix}=0.1$.

Other parameters (e.g., for the optimizer and triplet loss margin) are
adopted from FaceX-Zoo.
We run training on 8 NVIDIA RTX A5000 GPUs with a batch size of 4
per GPU (global batch size 32).
We run training for 8 epochs, completing it within 4 days.
As baselines, we compare \defense{} with \DOA, a defense tailored for
ad hoc semantic attacks using adversarial patches or eyeglasses (see
  \secref{sec:related}), DiffPure~\citep{nie2022diffusion}, a diffusion-based test-time purification defense, and standard filters considered in prior
work~\citep{le2025diffprivate}, including Gaussian blur, denoising via total
variation minimization, JPEG compression, feature squeezing, spatial
smoothing, and random noise injection.

\parheading{Computational Costs}
All preprocessing and training runs on 8 NVIDIA RTX A5000 GPUs.
GAN inversion (150 steps per image, $\sim$4.7 GPU-days) and PTI fine-tuning of the StyleGAN generator (125 steps, $\sim$4.3 GPU-days; reducible to $\sim$3.4 GPU-days at 100 steps) are one-time costs for the training set, shared across all training runs.
Unlike these preprocessing steps, adversarial training is not one-time: training each \facerec{} backbone takes approximately 4 days, and multiple runs were required for hyperparameter search, per standard practice for adversarially trained models.

\parheading{Metrics and \facerec{} Operating Point} 
We evaluate model robustness through accuracy (i.e., TPR) after
perturbing one of the samples from a positive pair.
Following standard practice, we calibrate the verification threshold
to meet a target FPR on clean data (without attacks).
Specifically, we perform the calibration on LFW's full (clean)
validation set (containing negative and positive pairs) to obtain an
FPR of 0.01, similar to~\cite{le2025diffprivate}.

\section{Experimental Results}
\label{sec:exp}

We now evaluate the \attack{} attack and \defense{} defense.

\subsection{\attack{} is Potent and Fast}
\label{subsec:semantic-risk}

\begin{table}[t!]
\centering
\begingroup
\small
\setlength{\tabcolsep}{4pt}
\begin{tabular}{llccccc}
\toprule
Dataset & Model & Clean & \multicolumn{4}{c}{\attack{} - budget $\beta$} \\
\cmidrule(lr){4-7}
& & & 1 & 1.5 & 2 & 3 \\
\midrule
\multirow{7}{*}{LFW}
& SwinT      & 99.1 & 94.9 & 88.0 & 78.7 & 52.6 \\
& LightCNN   & 99.1 & 90.3 & 84.7 & 72.7 & 40.5 \\
& MobileFace & 99.1 & 91.7 & 88.0 & 75.0 & 50.2 \\
& RepVGG     & 99.1 & 92.6 & 83.3 & 69.4 & 40.5 \\
& ResNet     & 99.1 & 93.5 & 83.8 & 70.8 & 39.5 \\
& MagFace    & 99.1 & 94.0 & 88.9 & 77.8 & 52.2 \\
& ArcFace    & 99.5 & 94.4 & 85.2 & 74.1 & 45.0 \\
\midrule
\multirow{7}{*}{VGG-Face}
& SwinT      & 100.0 & 85.3 & 80.8 & 65.4 & 45.5 \\
& LightCNN   & 98.7  & 84.0 & 72.4 & 60.3 & 34.0 \\
& MobileFace & 97.4  & 78.8 & 72.4 & 62.2 & 38.5 \\
& RepVGG     & 100.0 & 80.8 & 71.2 & 60.3 & 32.0 \\
& ResNet     & 100.0 & 84.0 & 72.4 & 59.6 & 34.6 \\
& MagFace    & 98.1  & 75.6 & 67.3 & 57.7 & 40.8 \\
& ArcFace    & 98.7  & 80.8 & 71.8 & 52.6 & 34.4 \\
\bottomrule
\end{tabular}
\endgroup
\caption{\textbf{White-box robustness against \attack{}}$^\dagger$ on LFW and VGG-Face.
We report benign accuracy and robust accuracy under increasing semantic attack budgets $\beta$; higher is better.
$^\dagger$95\% bootstrap CIs in \appref{app:stat_analysis}.}
\label{tab:wb_bs_fpr0p01}
\end{table}

\begin{table}[t!]
\centering
\begingroup
\small
\setlength{\tabcolsep}{3pt}
\begin{tabular}{llccccccc}
\toprule
Dataset & Model & Clean & \multicolumn{6}{c}{\diffprivate{} - budget $\lVert\Delta z\rVert$} \\
\cmidrule(lr){4-9}
& & & 1 & 2 & 3 & 4 & 5 & 6 \\
\midrule
\multirow{7}{*}{LFW}
& SwinT      & 99.1 & 99.5 & 98.6 & 98.6 & 85.6 & 50.5 & 49.1 \\
& LightCNN   & 99.1 & 99.1 & 97.7 & 89.4 & 62.0 & 32.9 & 28.7 \\
& MobileFace & 99.1 & 98.6 & 98.1 & 96.3 & 79.2 & 53.7 & 38.0 \\
& RepVGG     & 99.1 & 99.1 & 99.1 & 96.3 & 72.7 & 43.5 & 38.0 \\
& ResNet     & 99.1 & 99.1 & 98.6 & 93.5 & 71.8 & 41.7 & 33.8 \\
& MagFace    & 99.1 & 98.6 & 98.6 & 96.3 & 74.5 & 44.9 & 28.7 \\
& ArcFace    & 99.5 & 99.1 & 98.6 & 91.7 & 69.0 & 38.4 & 31.5 \\
\midrule
\multirow{7}{*}{VGG-Face}
& SwinT      & 100.0 & 83.9 & 76.8 & 64.5 & 41.9 & 26.5 & 19.4 \\
& LightCNN   & 98.7  & 69.7 & 58.7 & 45.8 & 25.2 & 18.1 & 14.8 \\
& MobileFace & 97.4  & 70.3 & 66.5 & 47.7 & 34.2 & 21.9 & 16.1 \\
& RepVGG     & 100.0 & 77.4 & 69.7 & 53.5 & 32.3 & 22.6 & 17.4 \\
& ResNet     & 100.0 & 77.4 & 67.7 & 51.0 & 30.3 & 22.6 & 17.4 \\
& MagFace    & 98.1  & 69.0 & 67.7 & 52.3 & 31.0 & 20.0 & 14.8 \\
& ArcFace    & 98.7  & 71.6 & 58.7 & 44.5 & 27.7 & 20.0 & 14.2 \\
\bottomrule
\end{tabular}
\endgroup
\caption{\textbf{White-box robustness against \diffprivate{}}$^\dagger$ on LFW and VGG-Face.
We report benign accuracy and robust accuracy under increasing latent perturbation norms $\lVert\Delta z\rVert$; higher is better.
$^\dagger$95\% bootstrap CIs in \appref{app:stat_analysis}.}
\label{tab:wb_dp_fpr0p01}
\end{table}

\parheading{White-box Attacks}
Tables~\ref{tab:wb_bs_fpr0p01} and~\ref{tab:wb_dp_fpr0p01} report the robust accuracy of the seven (undefended) \facerec{} backbones against \attack{} and \diffprivate{}, respectively, on the LFW and VGG-Face datasets when varying the attack budgets.
The results show that both attacks are potent in the white-box setting---despite high benign accuracy ($>$99\%), the robust accuracy drops as the attack budgets increase.
At \attack{}'s highest budget ($\beta{=}3$), it reduces average robust accuracy to $\sim$46\% on LFW and $\sim$37\% on VGG-Face.
At \diffprivate{} highest budget ($\lVert\Delta z\rVert{=}6$), it reduces average robust accuracy to $\sim$35.5\% on LFW and $\sim$16\% on VGG-Face.

\begin{figure}[t!]
\captionsetup{skip=1pt}
\centering
\setlength{\tabcolsep}{-4pt}
\footnotesize
\begin{tabular}{cc}
\includegraphics[width=0.55\columnwidth]{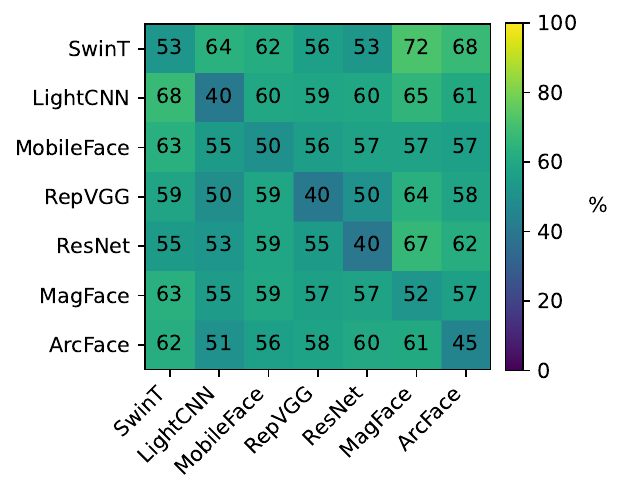} &
\includegraphics[width=0.55\columnwidth]{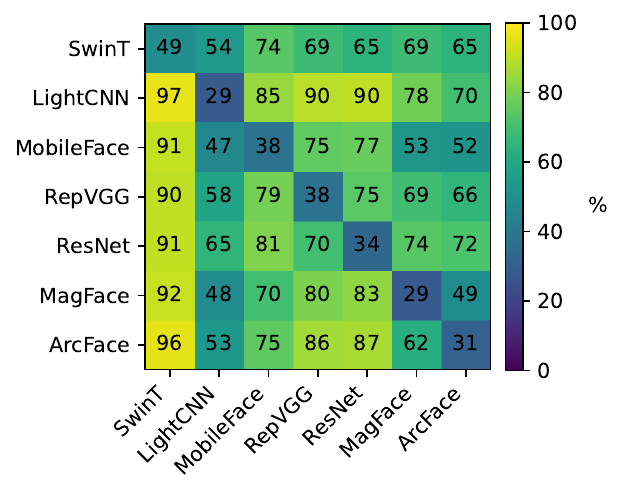} \\[-4pt]
{\scriptsize \textbf{(a)} LFW -- \attack{} ($\beta{=}3$)} &
{\scriptsize \textbf{(b)} LFW -- \diffprivate ($\lVert\Delta z\rVert{=}6$)} \\[2pt]
\includegraphics[width=0.55\columnwidth]{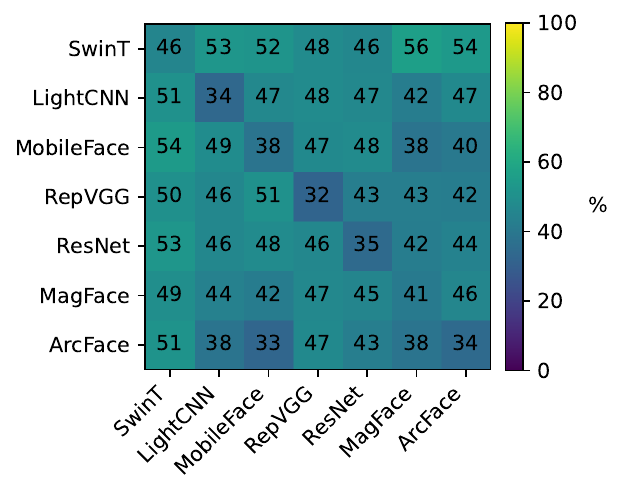} &
\includegraphics[width=0.55\columnwidth]{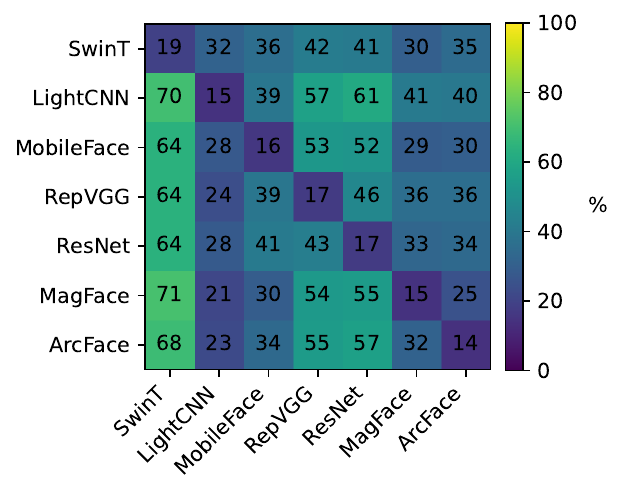} \\[-4pt]
{\scriptsize \textbf{(c)} VGG -- \attack{} ($\beta{=}3$)} &
{\scriptsize \textbf{(d)} VGG -- \diffprivate ($\lVert\Delta z\rVert{=}6$)} \\
\end{tabular}
\caption{\textbf{Robust accuracy of models against black-box attacks}.
    Each heatmap reports the robust accuracy of the model at the column, when transferring attacks from the model at the row. Diagonals correspond to white-box attacks.}
\label{fig:bb_fourup_fpr0p01}
\end{figure}

\parheading{Black-box Attacks}
\figref{fig:bb_fourup_fpr0p01} presents the robust accuracy of models when transferring attacks between models at the highest attack budgets considered.
In the heatmaps, off-diagonals show robust accuracy in black-box settings (transfer from the rows to columns), while diagonals correspond to white-box settings.
\emph{\attack{} exhibits strong transferability}---on both datasets, the heatmaps are near-uniform with off-diagonals typically within 5--15\% of the diagonal.
This indicates strong transferability, where attacks crafted on one model substantially reduce accuracy on others.

\diffprivate{} exhibits relatively weak transferability---off-diagonals are often 20--60\% higher than the diagonal, meaning attacks do not carry over well across architectures.
For instance, on LFW, evaluating on SwinT yields 49\% robust accuracy in the white-box setting compared to 90--97\% robust accuracy when transferring the attack from the convolutional networks to SwinT, suggesting that \diffprivate{} is architecture-specific.
To quantify this gap, we define a transferability score $T$ as the average ratio of black-box to white-box attack success rate across model pairs ($T{=}1$ indicates perfect transfer; see \appref{app:transferability} for the formal definition). Computed from \figref{fig:bb_fourup_fpr0p01}, \attack{} achieves $T{=}0.76$ (LFW) and $T{=}0.86$ (VGG-Face), while \diffprivate{} achieves $T{=}0.43$ (LFW) and $T{=}0.68$ (VGG-Face). We further show in \appref{app:transferability} that this gap is not an artifact of a specific parameter choice causing \diffprivate{} to overfit to the surrogate model: reducing \diffprivate{}'s iteration count does not improve its transferability, confirming that the gap reflects inherent differences between the two attacks.

\parheading{Attacks' Run Times}
We benchmark attacks' run times on an NVIDIA RTX A6000 GPU, executing each attack 100 times and averaging the run time.
For fair comparison, we use an equal batch size of 1 for both attacks.
Under this setting, \attack{} takes an average of 8{,}302 ms to complete per image compared to an average of 79{,}094 ms attained by \diffprivate{}, demonstrating approximately $\times$9.5 speedup.
This result highlights \attack{}'s better fit for adversarial training compared to other state-of-the-art attacks: \attack{} is not only highly effective in white-box settings and transferable in black-box settings, but is also significantly faster than \diffprivate{}.

\label{time_tradeoffs}
To further demonstrate that \attack{} attains superior time and success-rate trade-offs compared to \diffprivate{}, we execute both attacks under a matched wall-clock constraint on the same hardware. 
Specifically, we run the attacks on an NVIDIA A6000 GPU with a limit
of 6 seconds, which corresponds to approximately 30 iterations of
\attack{}. 
For \diffprivate{}, we provide a significant advantage by removing
caps on $\lVert\Delta z\rVert$ and the number of iterations,
constraining it solely by the run time.
\tabref{tab:time_tradeoff} lists the results.
It can be seen that, under equal execution time, \diffprivate{} leaves robust accuracy at 98.6\%, whereas \attack{} degrades it to 85.2--50.9\% (depending on $\beta$) against ResNet, on the LFW dataset. 
This result further highlights \attack{}'s efficiency and its
ability to attain high success rates within strict time constraints,
making it suitable for adversarial training.

\begin{table}[h]
\centering
\small
\begin{tabular*}{0.5\columnwidth}{@{\extracolsep{\fill}}llrr@{}}
\toprule
Attack & Budget & Time & Rob. Acc. \\
\midrule
\diffprivate{} & $\lVert\Delta z\rVert{=}\infty$ & 6s & 98.6\% \\
\attack{} & $\beta{=}1.5$ & 6s & 85.2\% \\
\attack{} & $\beta{=}2.0$ & 6s & 75.0\% \\
\attack{} & $\beta{=}3.0$ & 6s & 50.9\% \\
\bottomrule
\end{tabular*}
\caption{\textbf{Comparing attacks' success, on LFW and ResNet, under matched time constraints (6 seconds).}}
\label{tab:time_tradeoff}
\end{table}

\parheading{\attack{}'s Success Stems From Semantic Edits}
By construction, \attack{} cannot inject unconstrained pixel noise, as
its output is the StyleGAN3 decoding of an $\ell_2$-bounded latent
edit.
Four results indicate that its success is governed by semantic
changes:
\emph{(1)} $>$90\% of its perturbation energy lies below frequency
radius $r{\approx}15$ in Fourier domain vs.\ $r{\approx}30$ for \diffprivate{}
(\appref{imperceptible_perturbations});
\emph{(2)} noise-removal filters
barely affect \attack{} (${\le}6\%$ robust-accuracy increase) yet weaken \diffprivate{} by up to 36.6\% 
(see \secref{subsec:defenceeval});
\emph{(3)} adversarially training with pixel-space perturbations
shifts robustness against \attack{} by only ${\pm}{\approx}1\%$, but
against \diffprivate{} by up to ${+}9.3\%$ (\appref{app:ablation});
and
\emph{(4)} \attack{}'s principal attack directions decode to semantic
factors such as aging, pose, and illumination (\appref{app:pca}).


\subsection{\defense{} Improves Adversarial Robustness}
\label{subsec:defenceeval}

\begin{table}[t!]
\centering
\begingroup
\resizebox{\linewidth}{!}{
\setlength{\tabcolsep}{4pt}
\begin{tabular}{ll c c c c c c c c c c c}
\multicolumn{13}{c}{\textbf{LFW}}\\
\toprule
& & & \multicolumn{4}{c}{\texttt{\attack{}} - budget $\beta$} & \multicolumn{6}{c}{\texttt{DiffPrivate} - $\lVert\Delta z\rVert$} \\
\cmidrule(lr){4-7}\cmidrule(lr){8-13}
Backbone & Defense & Clean & 1 & 1.5 & 2 & 3 & 1 & 2 & 3 & 4 & 5 & 6 \\
\midrule
\multirow{3}{*}{ResNet}
  & No Defense & 99.1 & 93.5 & 83.8 & 70.8 & 39.5 & 99.1 & 98.6 & 93.5 & 71.8 & 41.7 & 33.8 \\
  & DOA        & 99.1 & 94.9 & 80.6 & 69.4 & 35.4 & 99.1 & 98.6 & 96.8 & 85.6 & \textbf{54.6} & 42.6 \\
  & \defense{} & \textbf{99.5} & \textbf{97.7} & \textbf{91.7} & \textbf{84.3} & \textbf{50.7} &
                 \textbf{99.5} & \textbf{99.5} & \textbf{97.7} & \textbf{86.6} & 54.2 & \textbf{47.2} \\
\midrule
\multirow{3}{*}{RepVGG}
  & No Defense & 99.1 & 92.6 & 83.3 & 69.4 & 40.5 & 99.1 & 99.1 & 96.3 & 72.7 & 43.5 & 38.0 \\
  & DOA        & \textbf{99.5} & 93.1 & 84.3 & 72.7 & 34.5 &
                 \textbf{99.5} & 99.1 & 97.7 & 89.4 & 59.7 & 43.1 \\
  & \defense{} & \textbf{99.5} & \textbf{97.2} & \textbf{92.1} & \textbf{84.7} & \textbf{54.3} &
                 \textbf{99.5} & \textbf{99.5} & \textbf{98.6} & \textbf{92.6} & \textbf{64.4} & \textbf{50.0} \\
\bottomrule
\end{tabular}
}
\vspace{6pt}

\resizebox{\linewidth}{!}{
\setlength{\tabcolsep}{4pt}
\begin{tabular}{ll c c c c c c c c c c c}
\multicolumn{13}{c}{\textbf{VGG-Face}}\\
\toprule
& & & \multicolumn{4}{c}{\texttt{\attack{}} - budget $\beta$} & \multicolumn{6}{c}{\texttt{DiffPrivate} - $\lVert\Delta z\rVert$} \\
\cmidrule(lr){4-7}\cmidrule(lr){8-13}
Backbone & Defense & Clean & 1 & 1.5 & 2 & 3 & 1 & 2 & 3 & 4 & 5 & 6 \\
\midrule
\multirow{3}{*}{ResNet}
  & No Defense & \textbf{100.0} & 84.0 & 72.4 & 59.6 & 34.6 &
                 77.4 & 67.7 & 51.0 & 30.3 & 22.6 & 17.4 \\
  & DOA        & 99.4 & 82.7 & 69.9 & 55.8 & 28.9 &
                 79.4 & 74.2 & 58.7 & 38.1 & 27.1 & 16.1 \\
  & \defense{} & \textbf{100.0} & \textbf{88.5} & \textbf{81.4} & \textbf{65.4} & \textbf{35.3} &
                 \textbf{81.3} & \textbf{76.1} & \textbf{63.2} & \textbf{41.9} & \textbf{27.7} & \textbf{18.1} \\
\midrule
\multirow{3}{*}{RepVGG}
  & No Defense & \textbf{100.0} & 80.8 & 71.2 & 60.3 & 32.0 &
                 77.4 & 69.7 & 53.5 & 32.3 & 22.6 & 17.4 \\
  & DOA        & 98.7 & 79.5 & 73.7 & 57.0 & 27.6 &
                 \textbf{81.3} & 74.2 & 57.4 & 40.0 & 26.5 & \textbf{21.3} \\
  & \defense{} & 99.4 & \textbf{89.1} & \textbf{80.8} & \textbf{69.9} & \textbf{35.7} &
                 80.6 & \textbf{76.1} & \textbf{62.6} & \textbf{43.2} & \textbf{31.0} & 20.6 \\
\bottomrule
\end{tabular}
}
\caption{\textbf{Evaluating defenses against white-box attacks}. For each defense and undefended model, we report benign accuracy and robust accuracy under varied attack budgets on the LFW and VGG-Face datasets, for both ResNet and RepVGG backbones.}
\label{tab:lfw_vgg_risk_fpr0p01}
\label{tab:def_bs_fpr0p01}
\label{tab:def_dp_fpr0p01}
\endgroup
\end{table}

\parheading{White-box Setting}
We execute white-box attacks against \defense{} and \DOA{} as well as against the undefended models (ResNet and RepVGG) at different attack budgets.
\tabref{tab:lfw_vgg_risk_fpr0p01} reports the results.
Compared to the undefended models, \defense{} shows a 0.4--20.9\% increase in robust accuracy on the different attack budgets against both \attack{} and \diffprivate{}, with the latter unseen during training.
The \DOA{} defense, tailored for ad hoc semantic attacks, shows a 0.4--0.7\% higher robust accuracy than \defense{} on a few attack budgets on \diffprivate{}, but otherwise trails behind \defense{}'s robust accuracy by 0.4--9.6\% against the \diffprivate{} attack.
However, against \attack{}, \DOA{} is counterproductive for most attack budgets, decreasing robust accuracy compared to the undefended model by up to 6\%.
Altogether, these results show that \defense{} reliably improves robust accuracy against general semantic attacks while generalizing to attacks unseen during training.
Importantly, \defense{} also maintains the benign accuracy on clean data as the undefended model or even roughly improves it.

\begin{figure*}[t!]
  \centering

  \begin{minipage}[t]{\textwidth}
    \centering
    \includegraphics[width=\textwidth]{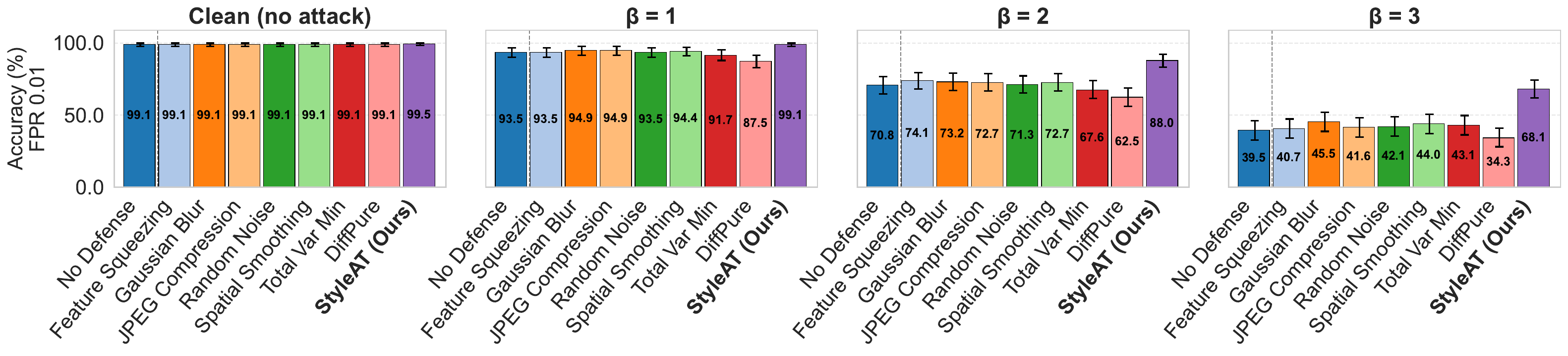}
    \par\vspace{2pt}{\small LFW}
  \end{minipage}

  \vspace{0.6em}

  \begin{minipage}[t]{\textwidth}
    \centering
    \includegraphics[width=\textwidth]{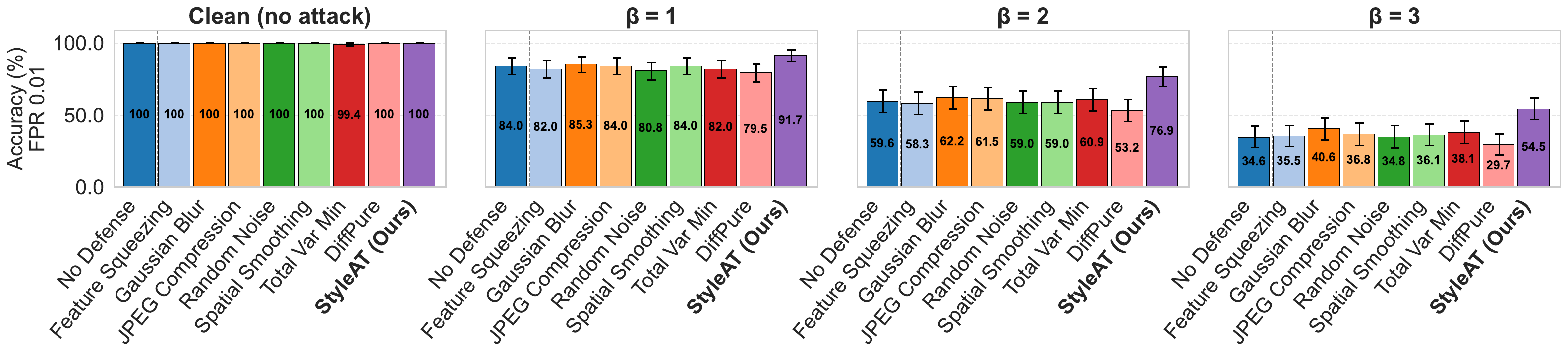}
    \par\vspace{2pt}{\small VGG-Face}
  \end{minipage}

  \caption{\textbf{Evaluating defenses against gray-box \attack{} attacks (ResNet), budgets $\beta \in \{1, 2, 3\}$.} The error bars present the 95\% bootstrap CIs. See \appref{app:bs_full} for all budgets}
  \label{fig:gbox-defenses-our-attack}
\end{figure*}

\begin{figure*}[t!]
  \centering

  \begin{minipage}[t]{\textwidth}
    \centering
    \includegraphics[width=\textwidth]{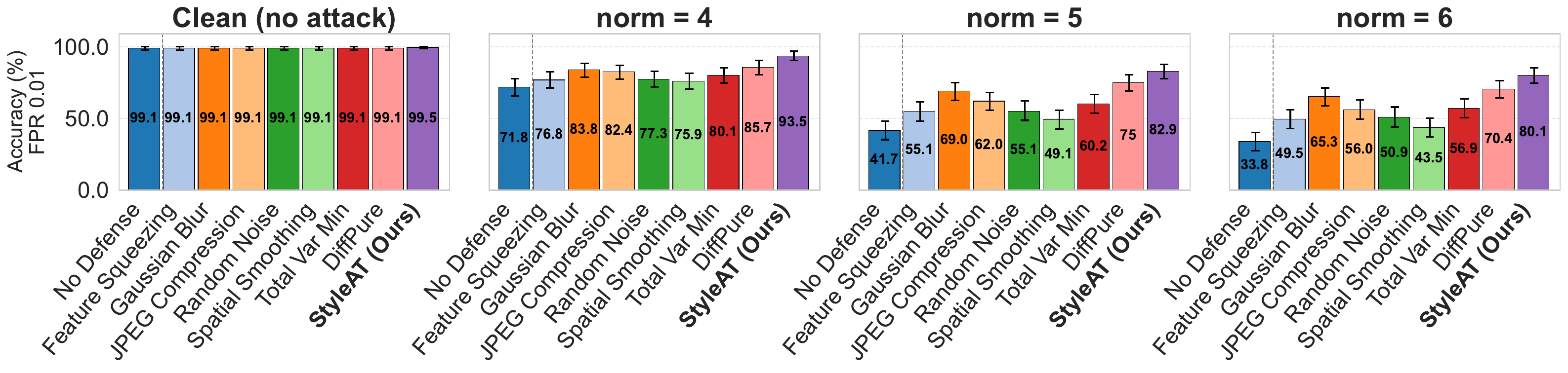}
    \par\vspace{2pt}{\small LFW}
  \end{minipage}

  \vspace{0.6em}

  \begin{minipage}[t]{\textwidth}
    \centering
    \includegraphics[width=\textwidth]{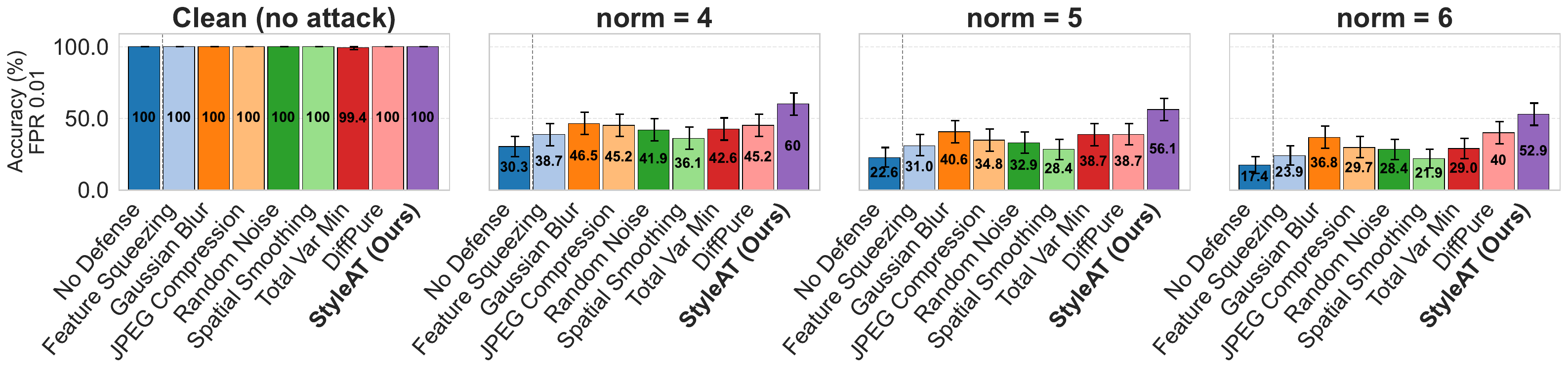}
    \par\vspace{2pt}{\small VGG-Face}
  \end{minipage}

  \caption{\textbf{Evaluating defenses against gray-box \diffprivate{} attacks (ResNet), norms 4--6.} The error bars present the 95\% bootstrap CIs. See \appref{app:dp_resnet_full} for all norms.}
  \label{fig:gbox-defenses-diffpriv}
\end{figure*}

\parheading{Gray-box Setting}
We also evaluate \defense{} and filter-based defenses against gray-box attacks, where the adversary produces attacks against the undefended models (ResNet and RepVGG), in a manner agnostic to the defense.
\figref{fig:gbox-defenses-our-attack} reports the robust accuracy achieved against \attack{} with the ResNet backbone (budgets $\beta \in \{1, 2, 3\}$ and clean; $\beta{=}1.5$ omitted for compactness, see \appref{app:bs_full}).
\figref{fig:gbox-defenses-diffpriv} reports \diffprivate{} results at norms $\lVert\Delta z\rVert{=}4\text{--}6$, where inter-defense differences are most pronounced; the full norm range appears in \appref{app:dp_resnet_full}.
Against \attack{}, it can be seen that filter-based defenses have
little impact on robustness, increasing robust accuracy by 6.0\% in
the best case compared to the undefended model.
\diffpure{} is in fact counterproductive against \attack{}, reducing robust accuracy by up to 6.0\% below the undefended baseline across all budgets and datasets.
In comparison, \defense{} results in up to 33.3\% increase in robust
accuracy.
The filter-based defenses are more useful against \diffprivate{},
increasing robust accuracy by up to 31.5\%.
By contrast, \diffpure{} substantially improves robustness against \diffprivate{}: it outperforms all filter-based defenses at moderate-to-high perturbation norms, improving robust accuracy by up to 36.6\% over the undefended model.
This asymmetry suggests that \diffpure{}'s purification is effective when adversarial structure aligns with latent diffusion dynamics, as in \diffprivate{}, but counterproductive against style-space manipulations.
Nonetheless, \defense{} outperforms all baselines, including \diffpure{}, against both attacks, with an increase of up to 50.9\% in robust accuracy over the undefended model against \diffprivate{}.
These results further highlight \defense{}'s utility against
defense-agnostic adversaries including against attacks not accounted
for during training (i.e., \diffprivate{}).

The results on the RepVGG backbone (Figs.~\ref{fig:gbox-defenses-our-attack-repvgg}--\ref{fig:gbox-defenses-diffpriv-repvgg} in \appref{app:styleat_repvgg}) are in line with the ResNet findings. \defense{} demonstrates superior robustness compared to filter-based defenses. For instance, against \diffprivate{} on the LFW dataset (at $\lVert\Delta z\rVert=6$), \defense{} achieves a robust accuracy of 88.9\%, significantly outperforming the strongest filter defense (Feature Squeezing at 69.9\%) and the undefended model (38.0\%). Similar trends are observed against the \attack{}, where filter-based defenses fail to provide meaningful robustness gains compared to \defense{}.

\section{Conclusion, Limitations, and Future Work}
\label{sec:conclusion}

Our work studies general semantic attacks against \facerec{}, proposing a new attack and a defense.
The proposed attack, \attack{}, produces high-fidelity images, attains high success rates, and is tunable, lending itself to being integrated into adversarial training.
We find that \attack{} achieves success rates on par with state-of-the-art attacks and even outperforms them in some settings, while being almost $\times$9.5 more time-efficient.
Our defense, \defense{}, is, to the best of our knowledge, the first defense tailored for general semantic attacks against \facerec{}. 
\defense{} builds on \attack{}, augmenting the training data with evasive samples produced via a weakened but efficient attack variant, leading to significant increases in robust accuracy in several settings we consider.
Notably, our findings also expose a limitation of generative semantic edits as a privacy-enhancing technology~\cite{le2024styleadv,le2025diffprivate}---\defense{}-hardened models still recognize faces protected with such edits.

\parheading{Limitations}
Our findings should be interpreted while taking several limitations into account.
First, we note that at the highest attack budgets ($\beta{=}3$ for \attack{} and $\lVert\Delta z\rVert{=}6$ for \diffprivate{}), human verification accuracy drops substantially (\appref{app:user_study}). Still, these settings should be interpreted as stress-tests probing worst-case vulnerabilities, not realistic deployment scenarios.
Second, due to computation constraints (evaluating all attack configurations against all models takes roughly one GPU-day per model), our evaluation is limited to 372 image pairs from LFW and VGG-Face. This scale is in line with prior evaluations of semantic attacks, including Adv-Attribute~\citep{jia2022adv} and Adv-Makeup~\citep{yin2021adv}, as robustness evaluation is bounded by attack-generation cost rather than by inference. While future work with larger compute budgets may evaluate on larger datasets (e.g., IJB-C or WebFace260M), the relatively tight confidence intervals we compute give us confidence that our findings will hold at larger evaluation scales.
Third, GAN inversion and PTI fine-tuning add an upfront one-time cost of approximately nine GPU-days for the training set, which may not suit all deployments.
Last, the improvements in robustness we demonstrate are purely empirical, without certified guarantees, and the models we train still exhibit some degree of susceptibility against general semantic attacks. Future work may seek to further increase adversarial robustness against general semantic attacks and derive theoretical guarantees, for instance through randomized smoothing~\citep{Cohen19RandSmooth} in the latent space.

\newpage
\section*{Acknowledgments}
This work has been supported in part
  by a grant from the Blavatnik Interdisciplinary Cyber Research Center (ICRC);
  by grant No.\ 2023641 from the United States-Israel Binational Science Foundation (BSF);
  by an Intel Rising Star Faculty Award;
  by Len Blavatnik and the Blavatnik Family foundation;
  by a Maof prize for outstanding young scientists;
  by the Ministry of Innovation, Science \& Technology, Israel (grant number 0603870071); and
  by a grant from the Tel Aviv University Center for AI and Data Science (TAD).
\bibliography{references}

\appendix
\onecolumn
\section{\defense{}'s Algorithm}
\label{app:defense_alg}

\algoref{alg:style_aware_at} presents the pseudocode of \defense{}.

\begin{algorithm}[t!]
\caption{\textsc{\defense{}} (per minibatch)}
\label{alg:style_aware_at}
\textbf{Input:} minibatch $\{(x_i, x_i^{+})\}_{i=1}^{B}$;  
StyleGAN encoder $E$ and generator $G$,  
feature extractor $f_\theta$,
latent attack steps $K$, step size $\alpha$, attack strength $\beta$, 
triplet loss margin $\mu$, 
pixel PGD steps $S$, $\alpha_{\mathrm{pix}}$ step, PGD radius $\varepsilon$; \\
\textbf{loss weights:} $\lambda_\text{Cln},\lambda_\text{AdvSem},\lambda_\text{AdvPix} \ge 0$ with $\lambda_\text{Cln} + \lambda_\text{AdvSem} + \lambda_\text{AdvPix} = 1$

\textbf{Output:} updated parameters $\theta$
\vspace{2pt}\hrule\vspace{4pt}
\begin{algorithmic}[1]\small
\FOR{$i=1$ \TO $B$}                               \label{line:inv_start}
    \STATE $l_i \leftarrow \text{OptimizeInv}\bigl(E(x_i), G\bigr)$  \textit{// Inversion (offline cacheable)}
\ENDFOR                                            \label{line:inv_end}
\FOR{$i=1$ \TO $B$}                               \label{line:atk_start}
        \STATE \textbf{Latent branch (\attack{}):}
        \STATE Sample unit direction $v_i \sim \mathcal{N}(0,I)$;  $v_i \leftarrow v_i/\|v_i\|_2$
        \STATE Sample $b_i \sim \mathcal{U}(0,\beta)$; \quad
        \STATE $\delta_i \leftarrow b_i\,v_i$
        \STATE $x_i \leftarrow \text{CropAlign}(x_i)$; \quad $x_i^{+} \leftarrow \text{CropAlign}(x_i^{+})$
        \FOR{$k=1$ \TO $K$}
        \STATE $x_i^{\text{adv}} \leftarrow \text{CropAlign}\bigl(G(l_i + \delta_i)\bigr)$
        \STATE $g \leftarrow \nabla_{\delta}\left[-\cos\bigl(f_\theta(x_i^{\text{adv}}),\,f_\theta(x_i)\bigr)\right]$ \textit{// Dodging: reduce cosine}
        \STATE $\delta_i \leftarrow \bigl(\delta_i + \alpha\,\frac{g}{||g||_\infty}\bigr)$ 
        \STATE Project $\delta_i \leftarrow \beta \cdot \frac{\delta_i}{\|\delta_i\|_2}$    
    \ENDFOR
    \STATE $x_i^{\text{adv}} \leftarrow \text{CropAlign}\bigl(G(l_i + \delta_i)\bigr)$
    \STATE \textbf{Pixel branch (PGD under $\ell_\infty(\varepsilon)$, \emph{around} $x_i^{\text{adv}}$):}
    \STATE $\hat{x_i}^{(0)} \leftarrow \mathrm{Clip}\big(x_i^{\text{adv}} + \mathcal{U}[-\varepsilon,\varepsilon]\big)$ \textit{// Random init near $x_i^{\text{adv}}$}
    \FOR{$s=1$ \TO $S$}
        \STATE $g_x \leftarrow \nabla_{x}\!\left[\cos\!\big( f_\theta(\hat{x_i}^{(s-1)}),\, f_\theta(x_i^{+})\big)\right]$
        \STATE $\hat{x_i}^{(s)} \leftarrow \hat{x_i}^{(s-1)} - \alpha_{\mathrm{pix}}\cdot \mathrm{sign}(g_x)$
        \STATE $\hat{x_i}^{(s)} \leftarrow \Pi_{B_\infty(x_i^{\text{adv}},\varepsilon)}\!\big(\hat{x}_i^{(s)}\big)$ \textit{// Project to $\ell_\infty$ ball around $x_i^{\text{adv}}$}
        \STATE $\hat{x_i}^{(s)} \leftarrow \mathrm{Clip}\big(\hat{x_i}^{(s)}\big)$ \textit{// valid pixel range}
    \ENDFOR
    \STATE $x_i^{\mathrm{pix}} \leftarrow \hat{x_i}^{(S)}$
        
\ENDFOR                                            \label{line:atk_end}
\FOR{$i=1$ \TO $B$} \label{line:embed_start}
    \STATE $e_i^{+}=f_\theta(x_i^{+})$; $e_i^{\text{cln}} = f_\theta(x_i)$;
    $e_i^{\text{advSem}}=f_\theta(x_i^{\text{adv}})$; $e_i^{\text{advPix}}=f_\theta(x_i^{\text{pix}})$
\ENDFOR \label{line:embed_end}
\STATE $L_{\text{Cln}} \leftarrow \mathrm{TripletLoss}\big( \{(e_i^{\text{cln}}, e_i^{+})\}_{i=1}^B, \mu,$
\STATE \hspace{2em} $\text{mine=batch-hardest-negative}\big)$
\STATE $L_{\text{AdvSem}} \leftarrow \mathrm{TripletLoss}\big( \{(e_i^{\text{advSem}}, e_i^{+})\}_{i=1}^B, \mu,$
\STATE \hspace{2em} $\text{mine=batch-hardest-negative}\big)$
\STATE $L_{\text{AdvPix}} \leftarrow \mathrm{TripletLoss}\big( \{(e_i^{\text{pix}}, e_i^{+})\}_{i=1}^B, \mu,$
\STATE \hspace{2em} $\text{mine=batch-hardest-negative}\big)$

\STATE $L \leftarrow \lambda_\text{Cln}\,L_{\text{Cln}} \;+\; \lambda_\text{AdvSem}\,L_{\text{AdvSem}} \;+\; \lambda_\text{AdvPix}\,L_{\text{AdvPix}}$ 
\STATE $\theta \leftarrow \theta - \eta\,\nabla_{\theta} L$
\STATE \textbf{return} $\theta$
\end{algorithmic}
\vspace{4pt}\hrule
\end{algorithm}

\section{Selection of Attacks' Perturbation Budgets}
\label{app:attack_budgets}                                                                                                                                                                                    

\begin{figure*}[t]
\centering
\setlength{\tabcolsep}{0pt}
\renewcommand{\arraystretch}{0.01}
\def\wBS{.10\linewidth}

\resizebox{0.60\linewidth}{!}{%
    \begin{tabular}{@{}ccccc@{}}
    \makebox[\wBS][c]{\scriptsize clean} & \makebox[\wBS][c]{\scriptsize 1} &
    \makebox[\wBS][c]{\scriptsize 1.5} & \makebox[\wBS][c]{\scriptsize 2} &
    \makebox[\wBS][c]{\scriptsize 3} \\[2pt]
    \includegraphics[width=\wBS]{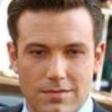} & \includegraphics[width=\wBS]{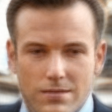} & \includegraphics[width=\wBS]{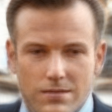} & \includegraphics[width=\wBS]{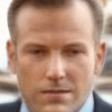} & \includegraphics[width=\wBS]{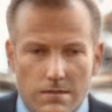} \\[-0.4pt]
    \includegraphics[width=\wBS]{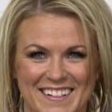} & \includegraphics[width=\wBS]{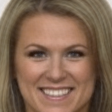} & \includegraphics[width=\wBS]{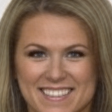} & \includegraphics[width=\wBS]{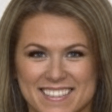} & \includegraphics[width=\wBS]{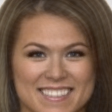} \\[-0.4pt]
    \includegraphics[width=\wBS]{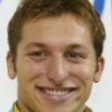} & \includegraphics[width=\wBS]{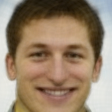} & \includegraphics[width=\wBS]{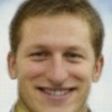} & \includegraphics[width=\wBS]{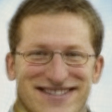} & \includegraphics[width=\wBS]{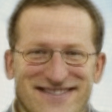} \\[-0.4pt]
    \includegraphics[width=\wBS]{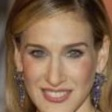} & \includegraphics[width=\wBS]{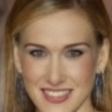} & \includegraphics[width=\wBS]{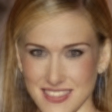} & \includegraphics[width=\wBS]{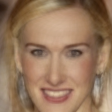} & \includegraphics[width=\wBS]{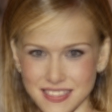} \\[-0.4pt]
    \end{tabular}%
}
\caption{Examples of edits produced by \attack{} when varying the attack budget $\beta\!\in\!\{1,1.5,2,3\}$.}
\label{fig:attack_examples}
\end{figure*}

\begin{figure*}[t]
\centering
\setlength{\tabcolsep}{0pt}
\renewcommand{\arraystretch}{0}
\resizebox{0.70\linewidth}{!}{
\newcommand{\wDP}{0.1\linewidth}
\begin{tabular}{@{}ccccccc@{}}
\makebox[\wDP][c]{\scriptsize clean} & \makebox[\wDP][c]{\scriptsize 1} & \makebox[\wDP][c]{\scriptsize 2} & \makebox[\wDP][c]{\scriptsize 3} & \makebox[\wDP][c]{\scriptsize 4} & \makebox[\wDP][c]{\scriptsize 5} & \makebox[\wDP][c]{\scriptsize 6} \\[2pt]
\includegraphics[width=\wDP]{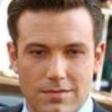} & \includegraphics[width=\wDP]{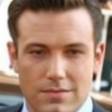} & \includegraphics[width=\wDP]{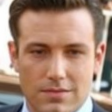} & \includegraphics[width=\wDP]{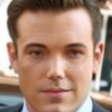} & \includegraphics[width=\wDP]{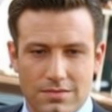} & \includegraphics[width=\wDP]{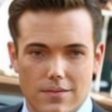} & \includegraphics[width=\wDP]{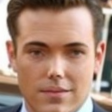} \\[-0.4pt]
\includegraphics[width=\wDP]{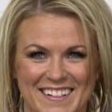} & \includegraphics[width=\wDP]{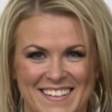} & \includegraphics[width=\wDP]{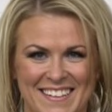} & \includegraphics[width=\wDP]{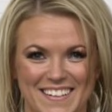} & \includegraphics[width=\wDP]{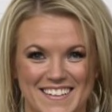} & \includegraphics[width=\wDP]{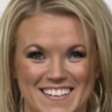} & \includegraphics[width=\wDP]{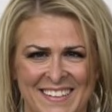} \\[-0.4pt]
\includegraphics[width=\wDP]{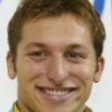} & \includegraphics[width=\wDP]{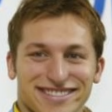} & \includegraphics[width=\wDP]{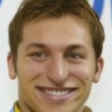} & \includegraphics[width=\wDP]{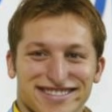} & \includegraphics[width=\wDP]{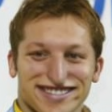} & \includegraphics[width=\wDP]{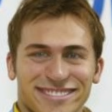} & \includegraphics[width=\wDP]{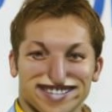} \\[-0.4pt]
\includegraphics[width=\wDP]{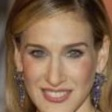} & \includegraphics[width=\wDP]{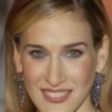} & \includegraphics[width=\wDP]{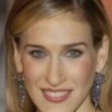} & \includegraphics[width=\wDP]{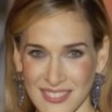} & \includegraphics[width=\wDP]{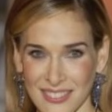} & \includegraphics[width=\wDP]{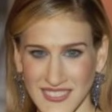} & \includegraphics[width=\wDP]{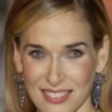} \\[-0.4pt]
\end{tabular}
}
\caption{Examples of edits produced by \diffprivate{} when varying the attack budget $\lVert\Delta z\rVert\!\in\!\{1,2,3,4,5,6\}$.}
\label{fig:diffprivate_examples}
\end{figure*}

\parheading{Qualitative Examples}
Figs.~\ref{fig:attack_examples}--\ref{fig:diffprivate_examples} provide qualitative examples of \attack{} and \diffprivate{} semantic attacks at different attack budgets. 
It can be seen that the original identities in the images become harder to identify as the attack budgets increase, leading to more aggressive semantic edits.

\begin{figure}[t]
    \centering
    \includegraphics[width=1\columnwidth]{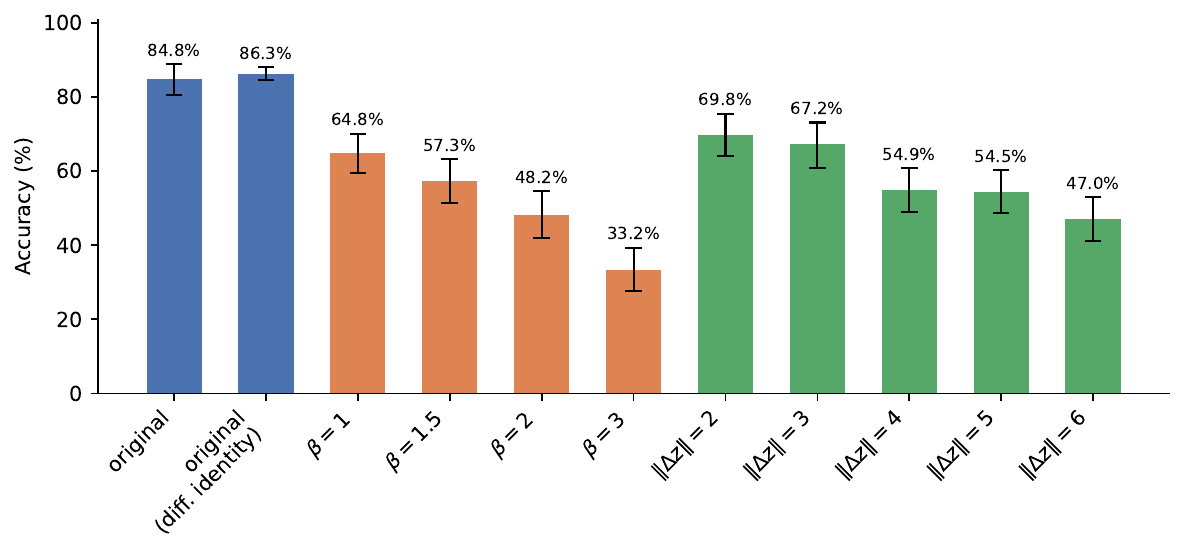}
    \caption{\textbf{Human verification accuracy under attack, with 95\% bootstrap confidence intervals.} Participants ($N=181$) evaluated image pairs at varying attack budgets. \textbf{Blue:} Original images (same and different identities); \textbf{Orange:} \attack{}; \textbf{Green:} \diffprivate{}.}
    \label{fig:user_study}
\end{figure}

\parheading{User Study: Humans' Verification Accuracy}
\label{subsec:user_study}
\label{app:user_study}
We conducted a user study with a convenience sample ($N=181$) to
evaluate human recognition performance under different attack budgets.  
Our participants were asked to verify 30 randomly sampled image
pairs each: 20 same-identity pairs where one image is clean and the
other is either clean or adversarially modified by \attack{}
or \diffprivate{}, and 10 different-identity pairs. The images were drawn at random from a combined pool of the LFW and VGG-Face datasets.
For each presented pair, participants needed to judge whether the
pairs depicted the same person (i.e., ``same'' or ``different'').
\figref{fig:user_study} details the results. 
We observe that verification accuracy declines smoothly as the attack
budgets ($\beta$ and $\lVert\Delta z\rVert$) increase, confirming
these knobs control edit strength.  
At lower attack budgets ($\beta$=1 and $\lVert\Delta z\rVert$=2),
participants maintain high accuracy (64.8\% and 69.8\%,
respectively), retaining approximately 76--82\% of the accuracy
achieved on original clean pairs ($\approx85\%$).  
Conversely, at the highest attack budgets we consider ($\beta$=3 and
$\lVert\Delta z\rVert$=6), accuracy drops significantly (33.2\% and
47\%, respectively), roughly 39--55\% of the clean images, justifying
the capping of $\beta\le$3 and $\lVert\Delta z\rVert\le$6.
The 95\% bootstrap CIs confirm the monotonically decreasing trend for \attack{}: $\beta{=}1$: 64.8\% [59.5--70.1\%], $\beta{=}1.5$: 57.3\% [51.5--63.1\%], $\beta{=}2$: 48.2\% [41.8--54.6\%], $\beta{=}3$: 33.2\% [27.5--39.3\%]. Consecutive budgets show at most minimal CI overlap; the CIs of $\beta{=}2$ and $\beta{=}3$ are fully non-overlapping (a $\sim$2.5pp gap between bounds), confirming that $\beta$ provides meaningful control over edit strength. For \diffprivate{}, the CIs reveal coarser budget sensitivity: $\lVert\Delta z\rVert{=}2$ and $\lVert\Delta z\rVert{=}3$ have heavily overlapping CIs (69.8\% [64.2--75.5\%] and 67.2\% [60.9--73.1\%]), as do $\lVert\Delta z\rVert{=}4$ and $\lVert\Delta z\rVert{=}5$ (54.9\% [49.0--60.8\%] and 54.5\% [48.7--60.2\%]); these two pairs are however non-overlapping with each other (the CIs of $\lVert\Delta z\rVert{=}3$ and $\lVert\Delta z\rVert{=}4$ just barely do not overlap), while $\lVert\Delta z\rVert{=}6$ (47.0\% [41.1--53.0\%]) overlaps partially with $\lVert\Delta z\rVert{=}4$ and $\lVert\Delta z\rVert{=}5$. The user study is thus a quantitative experiment, not merely a qualitative illustration.

\section{Ablation Study}
\label{app:ablation}

\parheading{Effect of $L_{\text{AdvPix}}$ During Training}
\tabref{tab:advpix_ablation_lfw} reports white-box robust accuracy on LFW for a ResNet model adversarially trained with and without $L_{\text{AdvPix}}$.
Against \diffprivate{}, enabling \(L_{\text{AdvPix}}\) boosts robust accuracy substantially for high attack budgets, by up to 9.3\%.
Against \attack{}, the effect of optimizing $L_{\text{AdvPix}}$ during training is minor and mixed, with -1.2--+0.93\% difference in robust accuracy in comparison to when $L_{\text{AdvPix}}$ is not optimized.
Overall, \(L_{\text{AdvPix}}\) primarily helps ameliorate \diffprivate{} at high attack budgets, while inducing only negligible changes against \attack{}. Thus, we include this term by default in our training objective.

A complementary ablation would train solely with pixel-space
adversarial training (i.e., without $L_{\text{AdvSem}}$).
Our evaluation contains a close proxy: \DOA{} -- adversarial training
with no semantic branch -- improves robustness against \diffprivate{}
(${+}12.9\%$ at $\lVert\Delta z\rVert{=}5$ on LFW) yet is
counterproductive against \attack{} (up to ${-}6\%$ at $\beta{=}3$;
\tabref{tab:lfw_vgg_risk_fpr0p01}).
Together with the ablation above, this indicates that
$\ell_\infty$-style robustness does not extend to large-norm,
low-frequency semantic edits: the pixel branch counters imperceptible
perturbations, while semantic robustness requires the latent branch.

\begin{table}[h]
\centering
\renewcommand{\arraystretch}{1.0}
\footnotesize

\textbf{BoundStyle ($\beta$)}\\[2pt]
\begin{tabular*}{\columnwidth}{@{\extracolsep{\fill}}lcccc@{}}
\toprule
 & 1 & 1.5 & 2 & 3 \\
\midrule
w/o $L_{\text{AdvPix}}$ & 98.61 & 90.74 & 83.80 & 51.87 \\
w/  $L_{\text{AdvPix}}$ & 97.69 & 91.67 & 84.26 & 50.72 \\
$\Delta$   & $-0.92$ & $+0.93$ & $+0.46$ & $-1.15$ \\
\bottomrule
\end{tabular*}

\vspace{6pt}
\textbf{DiffPrivate ($\|\Delta z\|$)}\\[2pt]
\begin{tabular*}{\columnwidth}{@{\extracolsep{\fill}}lcccccc@{}}
\toprule
 & 1 & 2 & 3 & 4 & 5 & 6 \\
\midrule
w/o $L_{\text{AdvPix}}$ & 99.5 & 99.1 & 97.7 & 79.6 & 46.3 & 38.0 \\
w/  $L_{\text{AdvPix}}$ & 99.5 & 99.5 & 97.7 & 86.6 & 54.2 & 47.2 \\
$\Delta$   & $0$ & $+0.5$ & $0$ & $+6.9$ & $+7.9$ & $+9.3$ \\
\bottomrule
\end{tabular*}
\caption{\textbf{The effect of $L_{\text{AdvPix}}$ on robust accuracy against white-box attacks.} We use LFW for evaluation.}
\label{tab:advpix_ablation_lfw}
\end{table}

\parheading{Attack Iterations}
\label{ablation_iterations}
We analyze the impact of the number of attack iterations on the
  effectiveness of attacks. 
\tabref{tab:ablation_iterations} shows the robust accuracy of the
ResNet model on LFW under \attack{} ($\beta=3$) and \diffprivate{}
($\|\Delta z\|=5$) for varying numbers of iterations. 
For both attacks, we observe that attack success saturates after a
certain number of steps.
For \attack{}, increasing iterations from 30 to 50 results in a
marginal accuracy drop of only 0.76\%. 
Similarly, for \diffprivate{}, extending the attack from 70 to 250
iterations yields a negligible decrease of 0.47\%. 
Based on these results, we fix the number of iterations to 30 for
\attack{} and 70 for \diffprivate{} to balance attack strength with
computational efficiency. 

\begin{table}[h]
\centering
\setlength{\tabcolsep}{6pt}
\renewcommand{\arraystretch}{1.1}
\footnotesize

\begin{minipage}{0.48\linewidth}
\centering
\textbf{BoundStyle ($\beta=3$)}\\[2pt]
\begin{tabular}{c c}
\toprule
Iterations & Accuracy (\%) \\
\midrule
3  & 51.16 \\
5  & 50.48 \\
10 & 47.44 \\
15 & 47.20 \\
\textbf{30} & \textbf{39.52} \\
50 & 38.76 \\
\bottomrule
\end{tabular}
\end{minipage}\hfill
\begin{minipage}{0.48\linewidth}
\centering
\textbf{DiffPrivate ($\|\Delta z\|=5$)}\\[2pt]
\begin{tabular}{c c}
\toprule
Iterations & Accuracy (\%) \\
\midrule
10 & 98.15 \\
50 & 51.85 \\
\textbf{70} & \textbf{41.67} \\
100 & 41.20 \\
250 & 41.20 \\
\bottomrule
\end{tabular}
\end{minipage}
\caption{\textbf{Ablation on attack iterations.} We report
    robust accuracy on LFW against \attack{} ($\beta=3$) and
    \diffprivate{} ($\|\Delta z\|=5$) with varying iteration
    counts. The selected number of iterations is in boldface.}
\label{tab:ablation_iterations}
\end{table}
\section{Imperceptible Perturbations in \diffprivate{}}
\label{imperceptible_perturbations}

\begin{figure}[t]
        \centering
        \includegraphics[width=\columnwidth]{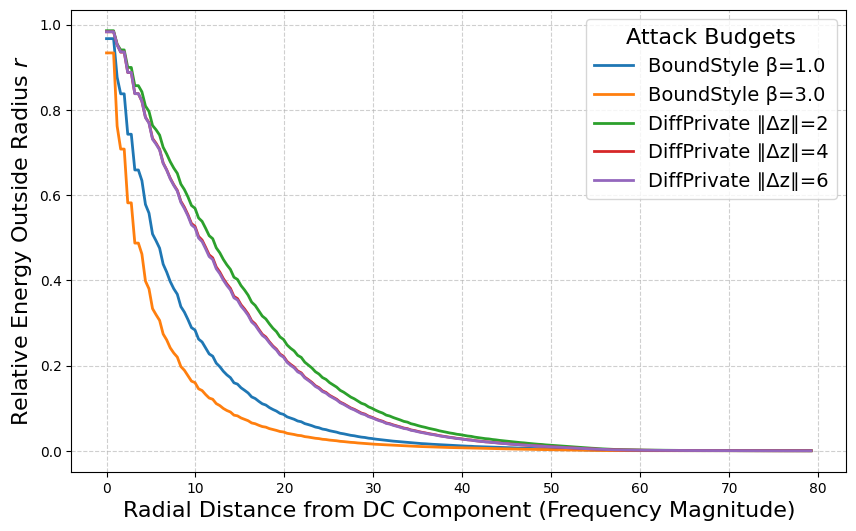}
        \caption{\textbf{Frequency energy decay of adversarial
    perturbations.} We measure the relative energy of the noise
    residual outside frequency radius $r$. \attack{} (orange/blue)
    decays rapidly, indicating changes are concentrated in low
    frequencies (semantic). \diffprivate{} (green/red/purple) decays
    slower, indicating a reliance on higher frequencies (imperceptible
    noise). Results are shown for the red channel of LFW, but we observe consistent behavior across all color channels.}
        \label{fig:energy_decay}
    \end{figure}

We now present additional evidence that \diffprivate{} introduces
imperceptible perturbations.
Recall that \emph{(1)} \diffprivate{}'s success rates decrease
significantly when applying semantics-preserving filters such as
JPEG compression (\figref{fig:gbox-defenses-diffpriv}
and \figref{fig:gbox-defenses-diffpriv-repvgg}); and
that \emph{(2)} adversarial training against
PGD in the pixel space (employed in \defense{}) improves
robustness against \diffprivate{}
(\tabref{tab:advpix_ablation_lfw}).
Both of these findings indicate that \diffprivate{} introduce
imperceptible perturbations besides semantic edits.
To further investigate imperceptible perturbations produced
by \diffprivate{}, we analyze the frequency-energy patterns of the
adversarial perturbations (i.e., the pixel-wise difference between
original and adversarial images).
To this end, we convert the perturbations produced by attacks (i.e., difference
between edited and clean images) to the Fourier domain and measure
the cumulative energy outside an increasing radial distance $r$
from the zero frequency (i.e., DC) component.
The result is monotonic curves decreasing from 1 to 0, as shown
in \figref{fig:energy_decay}, which illustrate the energy
distribution: a faster decay indicates energy concentrated in low
frequencies (i.e., semantic changes), while a slower decay implies
reliance on high frequencies.
We find that the decay is significantly faster for \attack{}
($>90\%$ of residual energy contained within $r \approx 15$) than
for \diffprivate{} ($>$90\% of residual energy within $r \approx
30$).
This results further demonstrates that \diffprivate{}'s success
can be partially attributed to non-semantic, imperceptible
perturbations.

\section{Additional Gray-box Defense Results}
\label{app:styleat_repvgg}

\begin{figure*}[h]
  \centering

  \begin{minipage}[t]{\textwidth}
    \centering
    \includegraphics[width=\textwidth]{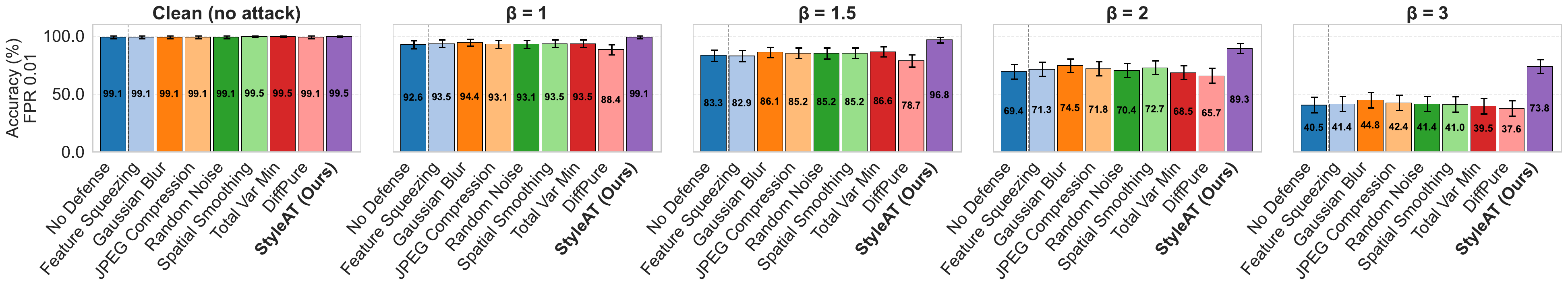}
    \par\vspace{2pt}{\small LFW}
  \end{minipage}

  \vspace{0.6em}

  \begin{minipage}[t]{\textwidth}
    \centering
    \includegraphics[width=\textwidth]{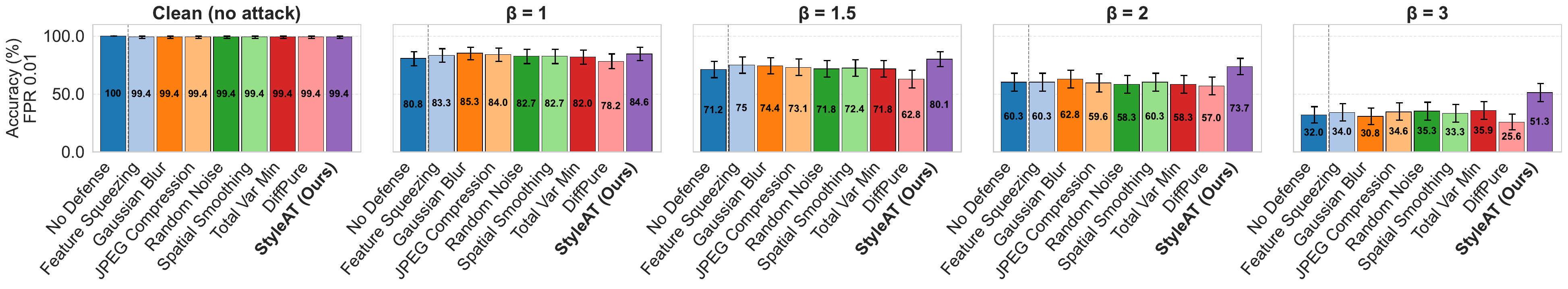}
    \par\vspace{2pt}{\small VGG-Face}
  \end{minipage}

  \caption{\textbf{Evaluating defenses against gray-box \attack{} attacks (RepVGG)}.}
  \label{fig:gbox-defenses-our-attack-repvgg}
\end{figure*}

\begin{figure*}[ht]
  \centering

  \begin{minipage}[t]{\textwidth}
    \centering
    \includegraphics[width=\textwidth]{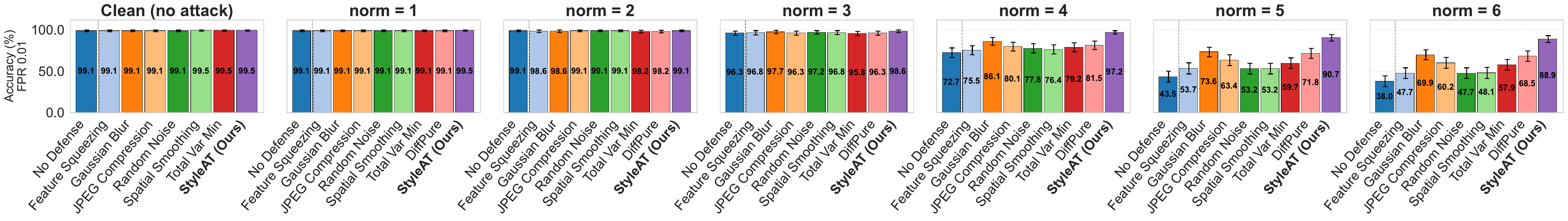}
    \par\vspace{2pt}{\small LFW}
  \end{minipage}

  \vspace{0.6em}

  \begin{minipage}[t]{\textwidth}
    \centering
    \includegraphics[width=\textwidth]{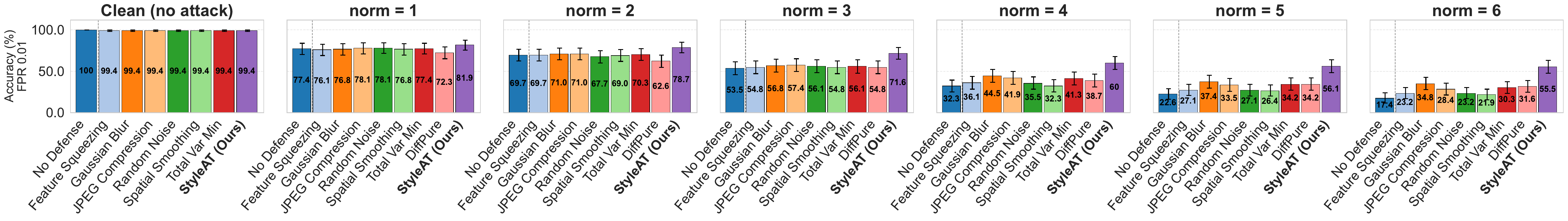}
    \par\vspace{2pt}{\small VGG-Face}
  \end{minipage}

  \caption{\textbf{Evaluating defenses against gray-box \diffprivate{} attacks (RepVGG)}.}
  \label{fig:gbox-defenses-diffpriv-repvgg}
\end{figure*}

\parheading{RepVGG backbone}
We present the gray-box defense evaluation results on the RepVGG
backbone in Figs.~\ref{fig:gbox-defenses-our-attack-repvgg}--\ref{fig:gbox-defenses-diffpriv-repvgg}.

\begin{figure*}[h]
  \centering

  \begin{minipage}[t]{\textwidth}
    \centering
    \includegraphics[width=\textwidth]{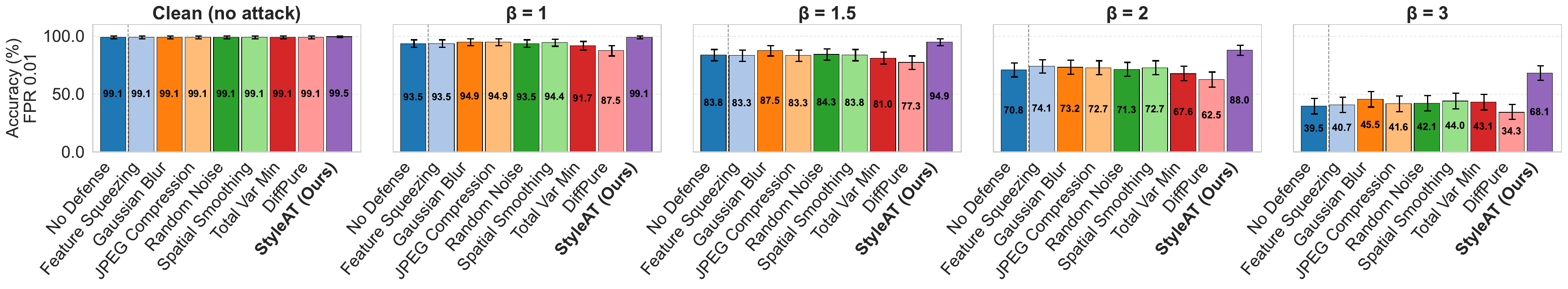}
    \par\vspace{2pt}{\small LFW}
  \end{minipage}

  \vspace{0.6em}

  \begin{minipage}[t]{\textwidth}
    \centering
    \includegraphics[width=\textwidth]{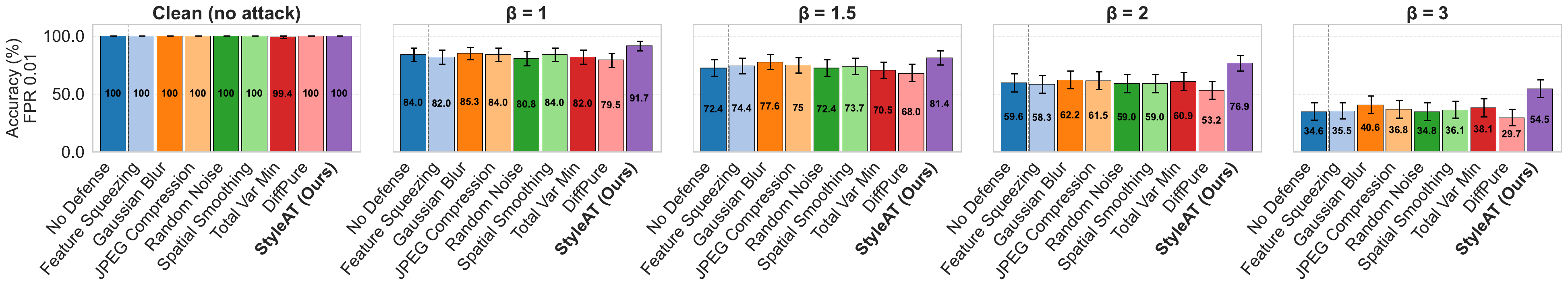}
    \par\vspace{2pt}{\small VGG-Face}
  \end{minipage}

  \caption{\textbf{Full \attack{} gray-box defense evaluation (ResNet, all budgets).} Robust accuracy (\%) across all attack budgets ($\beta \in \{1, 1.5, 2, 3\}$, plus clean). The error bars present the 95\% bootstrap CIs.}
  \label{fig:gbox-defenses-our-attack-full}
\end{figure*}

\begin{figure*}[h]
  \centering

  \begin{minipage}[t]{\textwidth}
    \centering
    \includegraphics[width=\textwidth]{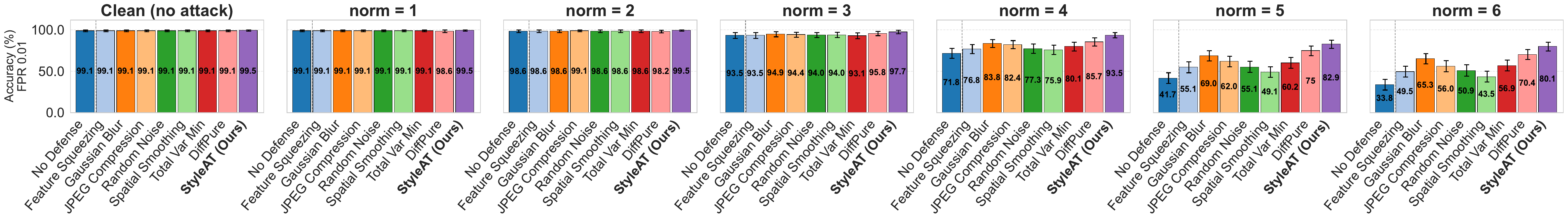}
    \par\vspace{2pt}{\small LFW}
  \end{minipage}

  \vspace{0.6em}

  \begin{minipage}[t]{\textwidth}
    \centering
    \includegraphics[width=\textwidth]{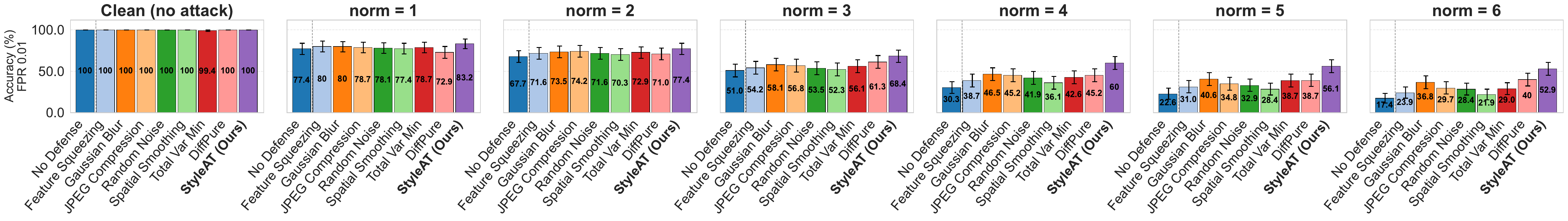}
    \par\vspace{2pt}{\small VGG-Face}
  \end{minipage}

  \caption{\textbf{Full \diffprivate{} gray-box defense evaluation (ResNet, all norms).} Robust accuracy (\%) across all perturbation norms ($\lVert\Delta z\rVert{=}1\text{--}6$). The error bars present the 95\% bootstrap CIs.}
  \label{fig:gbox-defenses-diffpriv-full}
\end{figure*}

\parheading{Full ResNet results}
\label{app:bs_full}
\label{app:dp_resnet_full}
We present the full gray-box defense evaluation results on the ResNet backbone in Figs.~\ref{fig:gbox-defenses-our-attack-full}--\ref{fig:gbox-defenses-diffpriv-full}.
The main-body figures omit $\beta{=}1.5$ (\figref{fig:gbox-defenses-our-attack}) and show only norms $\lVert\Delta z\rVert{=}4\text{--}6$ (\figref{fig:gbox-defenses-diffpriv}) for compactness; the full ranges confirm the same trends.

\section{Semantic Structure of \attack{} Adversarial Directions}
\label{app:pca}

To explore how \attack{} manipulates distinct
semantic attributes, we analyze the nature of the
adversarial perturbations.
To do so, we collect the latent perturbations $\delta$ generated
by \attack{} (at $\beta=3$) against the ResNet backbone on the LFW
dataset.
Subsequently, we then perform Principal Component Analysis (PCA) on
these perturbation vectors to identify the dominant directions of
variance in the attack space.

\begin{figure*}[t]
\centering
\setlength{\tabcolsep}{1pt}
\renewcommand{\arraystretch}{0.5}
\begin{tabular}{c@{\hspace{3pt}}cccccccccc}
& \multicolumn{2}{c}{\small Sample 1} & \multicolumn{2}{c}{\small Sample 2} & \multicolumn{2}{c}{\small Sample 3} & \multicolumn{2}{c}{\small Sample 4} & \multicolumn{2}{c}{\small Sample 5} \\[2pt]
& {\scriptsize Orig} & {\scriptsize Adv} & {\scriptsize Orig} & {\scriptsize Adv} & {\scriptsize Orig} & {\scriptsize Adv} & {\scriptsize Orig} & {\scriptsize Adv} & {\scriptsize Orig} & {\scriptsize Adv} \\[2pt]
\raisebox{0.033\textwidth}{\rotatebox{90}{\small\textbf{PC1}}} & \includegraphics[width=0.09\textwidth]{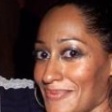} & \includegraphics[width=0.09\textwidth]{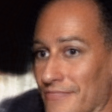} & \includegraphics[width=0.09\textwidth]{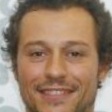} & \includegraphics[width=0.09\textwidth]{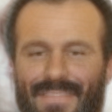} & \includegraphics[width=0.09\textwidth]{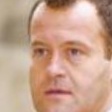} & \includegraphics[width=0.09\textwidth]{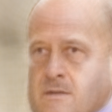} & \includegraphics[width=0.09\textwidth]{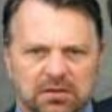} & \includegraphics[width=0.09\textwidth]{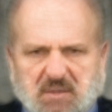} & \includegraphics[width=0.09\textwidth]{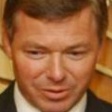} & \includegraphics[width=0.09\textwidth]{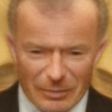} \\
\raisebox{0.033\textwidth}{\rotatebox{90}{\small\textbf{PC2}}} & \includegraphics[width=0.09\textwidth]{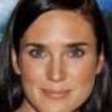} & \includegraphics[width=0.09\textwidth]{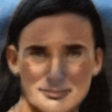} & \includegraphics[width=0.09\textwidth]{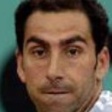} & \includegraphics[width=0.09\textwidth]{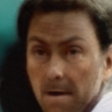} & \includegraphics[width=0.09\textwidth]{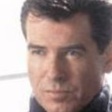} & \includegraphics[width=0.09\textwidth]{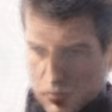} & \includegraphics[width=0.09\textwidth]{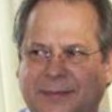} & \includegraphics[width=0.09\textwidth]{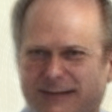} & \includegraphics[width=0.09\textwidth]{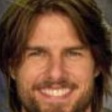} & \includegraphics[width=0.09\textwidth]{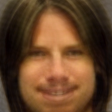} \\
\raisebox{0.033\textwidth}{\rotatebox{90}{\small\textbf{PC3}}} & \includegraphics[width=0.09\textwidth]{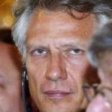} & \includegraphics[width=0.09\textwidth]{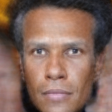} & \includegraphics[width=0.09\textwidth]{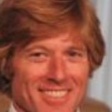} & \includegraphics[width=0.09\textwidth]{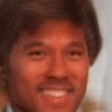} & \includegraphics[width=0.09\textwidth]{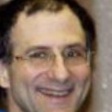} & \includegraphics[width=0.09\textwidth]{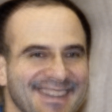} & \includegraphics[width=0.09\textwidth]{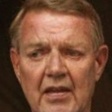} & \includegraphics[width=0.09\textwidth]{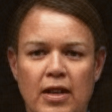} & \includegraphics[width=0.09\textwidth]{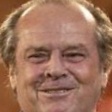} & \includegraphics[width=0.09\textwidth]{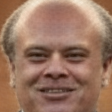} \\
\raisebox{0.033\textwidth}{\rotatebox{90}{\small\textbf{PC4}}} & \includegraphics[width=0.09\textwidth]{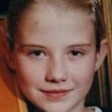} & \includegraphics[width=0.09\textwidth]{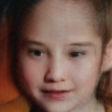} & \includegraphics[width=0.09\textwidth]{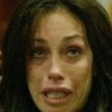} & \includegraphics[width=0.09\textwidth]{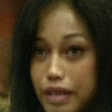} & \includegraphics[width=0.09\textwidth]{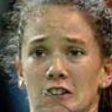} & \includegraphics[width=0.09\textwidth]{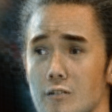} & \includegraphics[width=0.09\textwidth]{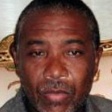} & \includegraphics[width=0.09\textwidth]{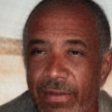} & \includegraphics[width=0.09\textwidth]{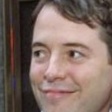} & \includegraphics[width=0.09\textwidth]{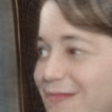} \\
\raisebox{0.033\textwidth}{\rotatebox{90}{\small\textbf{PC5}}} & \includegraphics[width=0.09\textwidth]{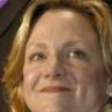} & \includegraphics[width=0.09\textwidth]{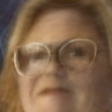} & \includegraphics[width=0.09\textwidth]{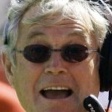} & \includegraphics[width=0.09\textwidth]{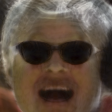} & \includegraphics[width=0.09\textwidth]{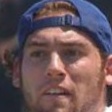} & \includegraphics[width=0.09\textwidth]{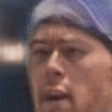} & \includegraphics[width=0.09\textwidth]{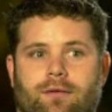} & \includegraphics[width=0.09\textwidth]{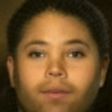} & \includegraphics[width=0.09\textwidth]{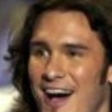} & \includegraphics[width=0.09\textwidth]{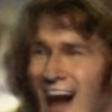} \\
\end{tabular}
\caption{Visualization of image samples whose adversarial perturbations align most strongly with the top five Principal Components (PCs) of the \attack{} perturbation distribution.
    Each column pair displays an original (left) and an adversarial (right) image. The dominant directions correspond to clear semantic attributes:
    \textbf{PC1} induces aging effects (older appearance);
    \textbf{PC2} alters nose shape and illumination;
    \textbf{PC3} creates a younger appearance;
    \textbf{PC4} corresponds to head pose adjustments; and
    \textbf{PC5} performs subtle structural changes to facial width.}
\label{fig:pca_pos_grid}
\end{figure*}

\figref{fig:pca_pos_grid} visualizes the top five principal
  components (PCs). 
  To interpret these abstract vectors, we project the perturbation
  $\delta_i$ of each test sample $i$ onto each PC. 
  Then, for every PC, we select the top five samples with the highest
  positive projection scores, i.e., the images whose adversarial
  manipulations align most strongly with that specific principal
  direction. 
  We visualize examples with $\beta$=3 to maximally emphasize the
  semantic nature of the directions. 
  Our visual analysis demonstrates that the dominant modes of the
  attack correspond to coherent semantic factors: 
  \textbf{PC1} captures aging (older appearance, thinning hair, and
  beard growth); 
  \textbf{PC2} modifies lighting and alters nose shape;
  \textbf{PC3} creates a younger appearance by smoothing skin texture
  and softening facial features; 
  \textbf{PC4} alters head pose; and
  \textbf{PC5} performs subtle structural changes to facial width.
  This analysis confirms that \attack{} discovers and exploits
  interpretable semantic weaknesses in the target \fr{} model. 

\section{Black-Box Transferability Analysis}
\label{app:transferability}

\parheading{Transferability Score}
We define the transferability score $T_i$ for source model $i$ as the average ratio of black-box attack success rate (ASR) to white-box ASR across all target models $j \neq i$:
\[
T_i = \frac{1}{N-1} \sum_{j \neq i} \frac{\mathrm{ASR}_{ij}}{\mathrm{ASR}_{ii}},
\qquad
T = \frac{1}{N} \sum_{i=1}^{N} T_i,
\]
where $\mathrm{ASR}_{ij} = 1 - \text{Rob. Acc.}_{ij}$ is the attack success rate when attacking model $i$ and evaluating on model $j$, and $N{=}7$ is the number of models.
$T{=}1$ indicates perfect transferability (black-box performance matches white-box) and lower values indicate weaker transfer.
\tabref{tab:T_summary} reports $T$ computed from the heatmaps in \figref{fig:bb_fourup_fpr0p01}.
Across both datasets, \attack{} consistently achieves higher $T$ than \diffprivate{}, confirming substantially stronger transferability.

\begin{table}[h]
\centering
\small
\setlength{\tabcolsep}{10pt}
\begin{tabular}{lcc}
\toprule
Attack & LFW & VGG-Face \\
\midrule
\attack{}      & 0.76 & 0.86 \\
\diffprivate{} & 0.43 & 0.68 \\
\bottomrule
\end{tabular}
\caption{Transferability score $T$ computed from the paper's black-box heatmaps (\figref{fig:bb_fourup_fpr0p01}).}
\label{tab:T_summary}
\end{table}

\parheading{Does Reducing DiffPrivate's Budget Improve Transferability?}
A possible concern is that the transferability gap reflects budget differences rather than inherent properties of the attacks: if optimizing for white-box success reduces transferability, a weaker (fewer-iteration) \diffprivate{} might transfer better.
We test this directly.
To isolate the effect of iteration count, we disable \diffprivate{}'s early stopping criterion (which halts optimization once the white-box model is fooled on the pair) and run it for a fixed number of iterations ranging from 20 to 70.
Note that this setup differs from the paper's main evaluation, which uses early stopping; the resulting $T$ values are therefore not directly comparable to those in \tabref{tab:T_summary}.

\tabref{tab:dp_transferability} reports the transferability score under each configuration.
Contrary to the hypothesis, $T$ does not decrease with more iterations---it is highest at 70 iterations, the setting that also maximizes white-box ASR.
This confirms that the transferability gap between \attack{} and \diffprivate{} reflects inherent differences in the two attack strategies rather than an artifact of attack-budget choices.

\begin{table}[h]
\centering
\small
\setlength{\tabcolsep}{8pt}
\begin{tabular}{cc}
\toprule
Iterations & $T$ \\
\midrule
20 & 0.533 \\
30 & 0.572 \\
40 & 0.541 \\
50 & 0.560 \\
60 & 0.577 \\
70 & 0.610 \\
\bottomrule
\end{tabular}
\caption{\textbf{\diffprivate{} transferability score $T$ across iteration counts on LFW} (early stopping disabled). $T$ does not decrease with more iterations, contradicting the hypothesis that stronger white-box attacks hurt black-box performance.}
\label{tab:dp_transferability}
\end{table}

\section{Bootstrap Confidence Intervals (CIs)}
\label{app:stat_analysis}

All confidence intervals are computed via the nonparametric percentile bootstrap~\citep{efron1979bootstrap} with 10{,}000 resamples, retaining 95\% coverage (2.5th and 97.5th percentiles). Each resample draws $n$ image pairs with replacement from the evaluation set and computes accuracy on the resampled set. The resulting CIs quantify sampling uncertainty due to the finite evaluation sets (LFW: $n{=}216$; VGG-Face: $n{=}156$). All values are reported as point estimate [\,lower--upper\,].

\parheading{Attacks Robustness CIs (Tables~\ref{tab:wb_bs_fpr0p01}--\ref{tab:wb_dp_fpr0p01})}
Tables~\ref{tab:ci_bs_lfw}--\ref{tab:ci_dp_vgg} report 95\% bootstrap CIs for all seven \facerec{} models on LFW and VGG-Face, for both \attack{} and \diffprivate{}.

\textit{\attack{}.} On LFW, the budget $\beta$ exerts strong, monotonically increasing attack pressure. Consecutive CI pairs are non-overlapping for all 7 models at $\beta{=}2{\to}3$, for 6/7 models at $\beta{=}1.5{\to}2$ (all except SwinT, which barely overlaps), and for 3/7 models (RepVGG, ResNet, ArcFace) already at $\beta{=}1{\to}1.5$. On VGG-Face, adjacent-budget separation is weaker at low budgets (all CIs overlap at $\beta{=}1{\to}1.5$, and only SwinT and ArcFace separate at $\beta{=}1.5{\to}2$), but the $\beta{=}2{\to}3$ transition is again fully non-overlapping across all 7 models.

\textit{\diffprivate{}.} On LFW, budget discrimination is strongest in the mid-range: all 7 models show non-overlapping CIs at both $\lVert\Delta z\rVert{=}3{\to}4$ and $\lVert\Delta z\rVert{=}4{\to}5$. At the low end, the $\lVert\Delta z\rVert{=}1{\to}2$ step yields no non-overlapping pairs and only 2/7 separate at $\lVert\Delta z\rVert{=}2{\to}3$, reflecting the limited attack gain at small perturbations. At the high end, the $\lVert\Delta z\rVert{=}5{\to}6$ step again shows mostly overlapping CIs (5/7 models), consistent with diminishing marginal gain at the highest budgets. On VGG-Face, the picture is more mixed: the $\lVert\Delta z\rVert{=}3{\to}4$ step separates for 6/7 models (similar to LFW), but discrimination collapses at higher budgets --- only SwinT separates at $\lVert\Delta z\rVert{=}4{\to}5$ and no model separates at $\lVert\Delta z\rVert{=}5{\to}6$. At the low end, no models separate at $\lVert\Delta z\rVert{=}1{\to}2$ and only 3/7 at $\lVert\Delta z\rVert{=}2{\to}3$. The wider CIs from the smaller VGG-Face evaluation set ($n{=}156$) likely account for the loss of discrimination at the high-budget end.

\begin{table}[h]
\centering
\begingroup
\scriptsize
\setlength{\tabcolsep}{3pt}
\begin{tabular}{lccccc}
\toprule
Model & Clean & $\beta{=}1$ & $\beta{=}1.5$ & $\beta{=}2$ & $\beta{=}3$ \\
\midrule
SwinT & 99.1 [97.7--100.0] & 94.9 [91.7--97.7] & 88.0 [83.3--92.1] & 78.7 [73.2--83.8] & 52.6 [46.0--59.5] \\
LightCNN & 99.1 [97.7--100.0] & 90.3 [86.1--94.0] & 84.7 [79.6--89.3] & 72.7 [66.7--78.7] & 40.5 [34.0--47.0] \\
MobileFace & 99.1 [97.7--100.0] & 91.7 [88.0--95.4] & 88.0 [83.3--92.1] & 75.0 [69.0--80.6] & 50.2 [43.5--57.0] \\
RepVGG & 99.1 [97.7--100.0] & 92.6 [88.9--95.8] & 83.3 [78.2--88.0] & 69.4 [63.0--75.5] & 40.5 [33.8--47.1] \\
ResNet & 99.1 [97.7--100.0] & 93.5 [89.8--96.8] & 83.8 [78.7--88.4] & 70.8 [64.8--76.8] & 39.5 [32.9--46.2] \\
MagFace & 99.1 [97.7--100.0] & 94.0 [90.7--96.8] & 88.9 [84.7--92.6] & 77.8 [72.2--83.3] & 52.2 [45.4--58.9] \\
ArcFace & 99.5 [98.6--100.0] & 94.4 [91.2--97.2] & 85.2 [80.6--89.8] & 74.1 [68.1--79.6] & 45.0 [38.3--51.7] \\
\bottomrule
\end{tabular}
\endgroup
\caption{\textbf{\attack{} CIs on LFW.} Robust accuracy (\%) [95\% bootstrap CI] under increasing $\beta$.}
\label{tab:ci_bs_lfw}
\end{table}

\begin{table}[h]
\centering
\begingroup
\scriptsize
\setlength{\tabcolsep}{3pt}
\begin{tabular}{lccccc}
\toprule
Model & Clean & $\beta{=}1$ & $\beta{=}1.5$ & $\beta{=}2$ & $\beta{=}3$ \\
\midrule
SwinT & 100.0 [100.0--100.0] & 85.3 [79.5--90.4] & 80.8 [74.4--86.5] & 65.4 [57.7--72.4] & 45.5 [37.8--53.2] \\
LightCNN & 98.7 [96.8--100.0] & 84.0 [78.2--89.7] & 72.4 [65.4--79.5] & 60.3 [52.6--68.0] & 34.0 [26.9--41.7] \\
MobileFace & 97.4 [94.9--99.4] & 78.8 [72.4--85.3] & 72.4 [65.4--78.8] & 62.2 [54.5--69.9] & 38.5 [30.8--46.1] \\
RepVGG & 100.0 [100.0--100.0] & 80.8 [74.4--87.2] & 71.2 [64.1--78.2] & 60.3 [52.6--68.0] & 32.0 [25.0--39.1] \\
ResNet & 100.0 [100.0--100.0] & 84.0 [78.2--89.7] & 72.4 [65.4--79.5] & 59.6 [51.9--67.3] & 34.6 [27.6--42.3] \\
MagFace & 98.1 [95.5--100.0] & 75.6 [68.6--82.0] & 67.3 [59.6--74.4] & 57.7 [50.0--65.4] & 40.8 [32.9--48.7] \\
ArcFace & 98.7 [96.8--100.0] & 80.8 [74.4--86.5] & 71.8 [64.7--78.8] & 52.6 [44.9--60.3] & 34.4 [27.3--42.2] \\
\bottomrule
\end{tabular}
\endgroup
\caption{\textbf{\attack{} CIs on VGG-Face.} Robust accuracy (\%) [95\% bootstrap CI] under increasing $\beta$.}
\label{tab:ci_bs_vgg}
\end{table}

\begin{table}[h]
\centering
\begingroup
\scriptsize
\setlength{\tabcolsep}{3pt}
\begin{tabular}{lcccccc}
\toprule
Model & $\lVert\Delta z\rVert{\!=\!}1$ & $\lVert\Delta z\rVert{\!=\!}2$ & $\lVert\Delta z\rVert{\!=\!}3$ & $\lVert\Delta z\rVert{\!=\!}4$ & $\lVert\Delta z\rVert{\!=\!}5$ & $\lVert\Delta z\rVert{\!=\!}6$ \\
\midrule
SwinT & 99.5 [98.6--100.0] & 98.6 [96.8--100.0] & 98.6 [96.8--100.0] & 85.7 [81.0--90.3] & 50.5 [43.5--57.4] & 49.1 [42.6--55.6] \\
LightCNN & 99.1 [97.7--100.0] & 97.7 [95.4--99.5] & 89.3 [85.2--93.5] & 62.0 [55.6--68.5] & 32.9 [26.9--39.4] & 28.7 [22.7--34.7] \\
MobileFace & 98.6 [96.8--100.0] & 98.2 [96.3--99.5] & 96.3 [93.5--98.6] & 79.2 [73.6--84.7] & 53.7 [46.8--60.2] & 38.0 [31.5--44.4] \\
RepVGG & 99.1 [97.7--100.0] & 99.1 [97.7--100.0] & 96.3 [93.5--98.6] & 72.7 [66.7--78.7] & 43.5 [37.0--50.5] & 38.0 [31.5--44.4] \\
ResNet & 99.1 [97.7--100.0] & 98.6 [96.8--100.0] & 93.5 [90.3--96.8] & 71.8 [65.7--77.8] & 41.7 [35.2--48.6] & 33.8 [27.8--40.3] \\
MagFace & 98.6 [96.8--100.0] & 98.6 [96.8--100.0] & 96.3 [93.5--98.6] & 74.5 [68.5--80.1] & 44.9 [38.4--51.9] & 28.7 [22.7--34.7] \\
ArcFace & 99.1 [97.7--100.0] & 98.6 [96.8--100.0] & 91.7 [88.0--95.4] & 69.0 [62.5--75.0] & 38.4 [31.9--44.9] & 31.5 [25.5--37.5] \\
\bottomrule
\end{tabular}
\endgroup
\caption{\textbf{\diffprivate{} CIs on LFW.} Robust accuracy (\%) [95\% bootstrap CI] under increasing $\lVert\Delta z\rVert$.}
\label{tab:ci_dp_lfw}
\end{table}

\begin{table}[h]
\centering
\begingroup
\scriptsize
\setlength{\tabcolsep}{3pt}
\begin{tabular}{lcccccc}
\toprule
Model & $\lVert\Delta z\rVert{\!=\!}1$ & $\lVert\Delta z\rVert{\!=\!}2$ & $\lVert\Delta z\rVert{\!=\!}3$ & $\lVert\Delta z\rVert{\!=\!}4$ & $\lVert\Delta z\rVert{\!=\!}5$ & $\lVert\Delta z\rVert{\!=\!}6$ \\
\midrule
SwinT & 83.9 [78.1--89.0] & 76.8 [69.7--83.2] & 64.5 [56.8--71.6] & 41.9 [34.2--49.7] & 26.4 [19.4--33.5] & 19.4 [13.6--25.8] \\
LightCNN & 69.7 [62.6--76.8] & 58.7 [51.0--66.5] & 45.8 [38.1--53.5] & 25.2 [18.7--32.3] & 18.1 [12.3--24.5] & 14.8 [9.7--20.6] \\
MobileFace & 70.3 [63.2--77.4] & 66.5 [58.7--73.5] & 47.7 [40.0--55.5] & 34.2 [27.1--41.9] & 21.9 [15.5--28.4] & 16.1 [10.3--21.9] \\
RepVGG & 77.4 [70.3--83.9] & 69.7 [62.6--76.8] & 53.5 [45.8--61.3] & 32.3 [25.2--39.4] & 22.6 [16.1--29.0] & 17.4 [11.6--23.9] \\
ResNet & 77.4 [71.0--83.9] & 67.7 [60.0--74.8] & 51.0 [43.2--58.7] & 30.3 [23.2--37.4] & 22.6 [16.1--29.0] & 17.4 [11.6--23.9] \\
MagFace & 69.0 [61.9--76.1] & 67.7 [60.0--74.8] & 52.3 [44.5--60.0] & 31.0 [23.9--38.1] & 20.0 [14.2--26.4] & 14.8 [9.7--20.6] \\
ArcFace & 71.6 [64.5--78.7] & 58.7 [51.0--66.5] & 44.5 [36.8--52.3] & 27.7 [20.6--34.8] & 20.0 [14.2--26.4] & 14.2 [9.0--20.0] \\
\bottomrule
\end{tabular}
\endgroup
\caption{\textbf{\diffprivate{} CIs on VGG-Face.} Robust accuracy (\%) [95\% bootstrap CI] under increasing $\lVert\Delta z\rVert$.}
\label{tab:ci_dp_vgg}
\end{table}

\begin{figure*}[h]
  \centering
  \begin{minipage}[t]{0.5\textwidth}
    \includegraphics[width=\textwidth]{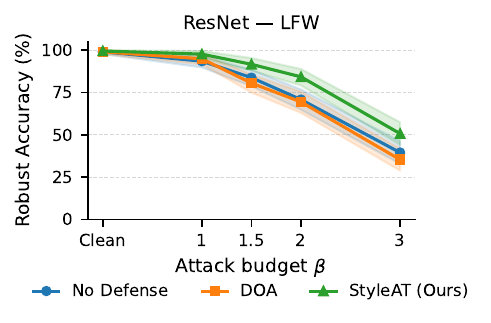}
  \end{minipage}\hfill
  \begin{minipage}[t]{0.5\textwidth}
    \includegraphics[width=\textwidth]{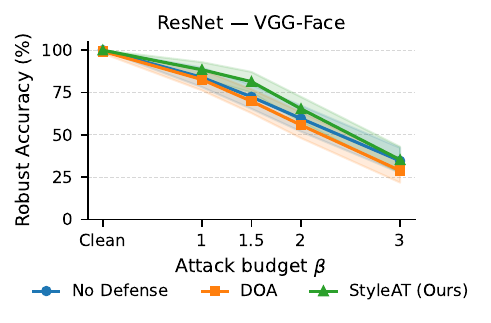}
  \end{minipage}

  \begin{minipage}[t]{0.5\textwidth}
    \includegraphics[width=\textwidth]{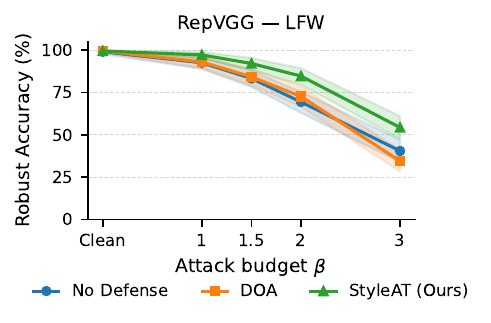}
  \end{minipage}\hfill
  \begin{minipage}[t]{0.5\textwidth}
    \includegraphics[width=\textwidth]{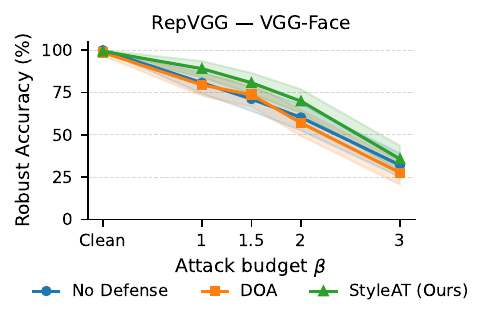}
  \end{minipage}

  \caption{\textbf{Defense robustness with 95\% bootstrap CIs --- \attack{}.}
    Each plot shows robust accuracy (\%) vs.\ budget $\beta$ for No Defense, DOA, and \defense{} (ours), with shaded CI bands.
    Left column: LFW; right column: VGG-Face.
    Top row: ResNet; bottom row: RepVGG.}
  \label{fig:ci_defense_bs}
\end{figure*}

\begin{figure*}[h]
  \centering
  \begin{minipage}[t]{0.5\textwidth}
    \includegraphics[width=\textwidth]{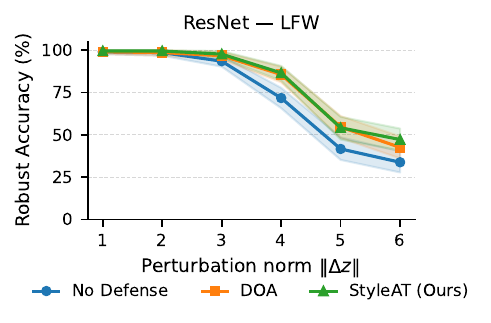}
  \end{minipage}\hfill
  \begin{minipage}[t]{0.5\textwidth}
    \includegraphics[width=\textwidth]{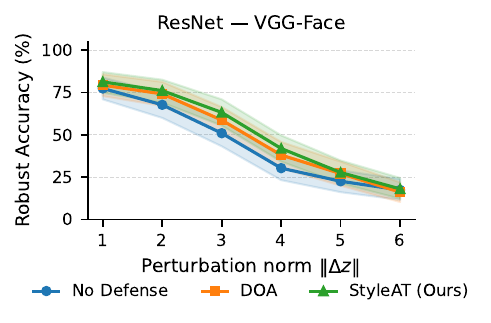}
  \end{minipage}

  \begin{minipage}[t]{0.5\textwidth}
    \includegraphics[width=\textwidth]{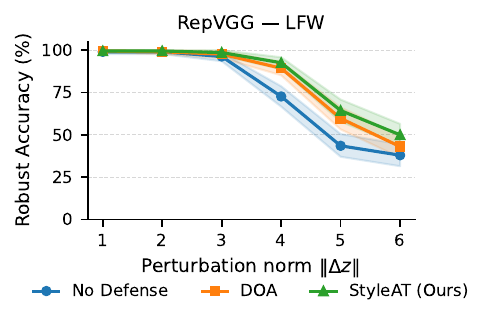}
  \end{minipage}\hfill
  \begin{minipage}[t]{0.5\textwidth}
    \includegraphics[width=\textwidth]{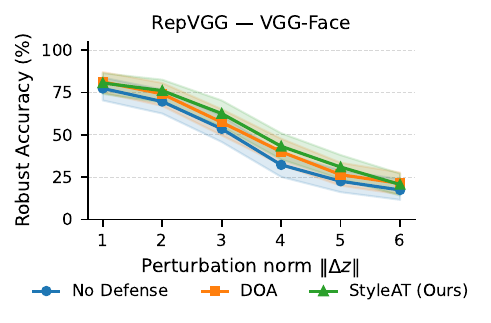}
  \end{minipage}

  \caption{\textbf{Defense robustness with 95\% bootstrap CIs --- \diffprivate{}.}
    Each plot shows robust accuracy (\%) vs.\ perturbation norm $\lVert\Delta z\rVert$ for No Defense, DOA, and \defense{} (ours), with shaded CI bands.
    Left column: LFW; right column: VGG-Face.
    Top row: ResNet; bottom row: RepVGG.}
  \label{fig:ci_defense_dp}
\end{figure*}

\parheading{Defense Robustness CIs (Table~\ref{tab:lfw_vgg_risk_fpr0p01})}
Figs.~\ref{fig:ci_defense_bs}--\ref{fig:ci_defense_dp} show 95\% bootstrap CI bands for the white-box defense evaluation across both backbones, datasets, and attacks.

\textit{\attack{}.}
On LFW, \defense{} is statistically separated from both No Defense and DOA across moderate-to-high budgets ($\beta{=}1.5$ through $\beta{=}3$), confirming that the robustness gains are reliable and not a sampling artifact.
At the lowest budget ($\beta{=}1$), all three defenses overlap, consistent with limited attack pressure.
On VGG-Face, CIs overlap throughout, though the overlap is generally small --- particularly against No Defense --- reflecting a consistent point-estimate advantage for \defense{} that falls short of full statistical separation owing to the wider intervals from the smaller evaluation set ($n{=}156$).

\textit{\diffprivate{}.}
On LFW, \defense{} is clearly separated from No Defense at high budgets ($\lVert\Delta z\rVert{\ge}4$), where the attack is most damaging.
Against DOA, CIs overlap throughout, reflecting that the two defenses achieve similar robust accuracy under \diffprivate{}.
On VGG-Face, CIs overlap throughout against both baselines, though not fully - the point estimates consistently favor \defense{}, particularly over No Defense, across all budgets.

\clearpage  
\end{document}